\documentclass[review]{elsarticle}

\usepackage{lineno, hyperref, amsmath, multirow, epstopdf}

\usepackage{amssymb}		
\usepackage{subfigure}		
\usepackage[utf8]{inputenc}
\usepackage{url}
\usepackage{mdframed}

\modulolinenumbers[5]

\journal{Robotics and Autonomous Systems}

\begin{document}

\begin{frontmatter}

\title{On the Manipulation of Articulated Objects in Human-Robot Cooperation Scenarios}

\author{Alessio Capitanelli}
\author{Marco Maratea}

\author{Fulvio Mastrogiovanni\fnref{mark:corresponding}}
\fntext[mark:corresponding]{Corresponding author. \textit{Email address}: \texttt{fulvio.mastrogiovanni@unige.it}}

\author{Mauro Vallati}

\address{A. Capitanelli, M. Maratea and F. Mastrogiovanni are with the Department of Informatics, Bioengineering, Robotics and Systems Engineering, University of Genoa, via Opera Pia 13, 16145 Genoa, Italy.
\\M. Vallati is with the School of Computing and Engineering, University of Huddersfield, Queensgate, HD13DH, Huddersfield, United Kingdom.}

\begin{abstract}

Articulated and flexible objects constitute a challenge for robot manipulation tasks, but are present in different real-world settings, including home and industrial environments.
Current approaches to the manipulation of articulated and flexible objects employ \textit{ad hoc} strategies to sequence and perform actions on them depending on a number of physical or geometrical characteristics related to those objects, as well as on an \textit{a priori} classification of target object configurations. 

In this paper, we propose an action planning and execution framework, which (i) considers abstract representations of articulated or flexible objects, (ii) integrates action planning to reason upon such configurations and to sequence an appropriate set of actions with the aim of obtaining a target configuration provided as a goal, and (iii) is able to cooperate with humans to collaboratively carry out the plan.

On the one hand, we show that a trade-off exists between the way articulated or flexible objects are perceived and how the system represents them.
Such a trade-off greatly impacts on the complexity of the planning process.
On the other hand, we demonstrate the system's capabilities in allowing humans to interrupt robot action execution, and -- in general -- to contribute to the whole manipulation process. 

Results related to planning performance are discussed, and examples with a Baxter dual-arm manipulator performing actions collaboratively with humans are shown.

\end{abstract}


\end{frontmatter}


\section{Introduction}
\label{sec:introduction}

The introduction of the Industry 4.0 paradigm is expected to redefine the nature of shop-floor environments along many directions, including the role played by robots in the manufacturing process \cite{Krugeretal2009, Heyer2010}.
One of the main tenets considered in Industry 4.0 is the increased customer satisfaction via a high degree of product personalization and just-in-time delivery. 
On the one hand, a higher level of flexibility in manufacturing processes is needed to cope with such diversified demands, especially in low-automation tasks.
On the other hand, skilful robots working alongside humans can be regarded as a valuable aid to shop-floor operators, who can supervise robots' work and intervene when needed \cite{Tsarouchietal2016}, whereas robots can be tasked with difficult or otherwise stressful operations.

Human-robot cooperation (HRC) processes in shop-floor environments are a specific form of human-robot interaction (HRI) with at least two important specificities.
The first is related to the fact that the cooperation is targeted towards a well-defined objective (e.g., an assemblage, a unit test, a cable harnessing operation), which must be typically achieved in a short amount of time. 
The second has to do with the fact that humans need to feel (at least partially) in control \cite{Baragliaetal2016, Munzeretal2017}:
although grounded in the cooperation process, their behaviours could be unpredictable in specific cases, with obvious concerns about their safety \cite{Deneietal2015, HaddadinCroft2016}; they may not fully understand robot goals \cite{Chakrabortietal2017}; robot actions may not be considered appropriate for the peculiar cooperation objectives \cite{GoodrichSchultz2007, Munzeretal2017}.

As far as the cooperation process is concerned, two high-level directives must be taken into account:
\begin{enumerate}
\item[$D_1$] \textit{cooperation models} (and robot action planning techniques) enforcing the prescribed objectives must be adopted \cite{JohannsmeierHaddadin2017, Darvishetal2017};
\item[$D_2$] the robot must be flexible enough to adapt to human operator actions avoiding a purely reactive approach \cite{Dautenhahn2007, Prewettetal2010}, and to make its intentions clear \cite{ClairMataric2015, Ronconeetal2017}.  
\end{enumerate}
These two directives lead to three functional requirements for a HRC architecture.
The robot must be able to:
\begin{enumerate}
\item[$R_1$] (at least implicitly) recognize the effects of human operator actions \cite{LiuWang2017};
\item[$R_2$] adapt its behaviour on the basis of two elements: human operator actions themselves and the whole cooperation objectives;
\item[$R_3$] employ planning techniques allowing for a fast action re-planning when needed, e.g., when planned actions cannot be executed for sudden changes in the environment or inaccurate modelling assumptions \cite{Srivastavaetal2014}.  
\end{enumerate}

\begin{figure}[t!]
\centering
\begin{subfigure}{}
\centering
\includegraphics[width = 2.1in]{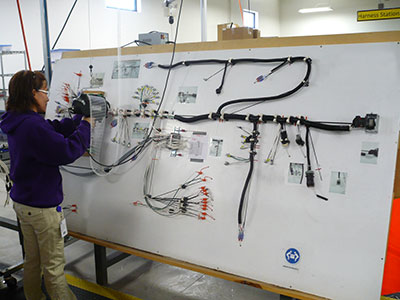}
\end{subfigure}
 \begin{subfigure}{}
\centering
\includegraphics[width = 2.36in]{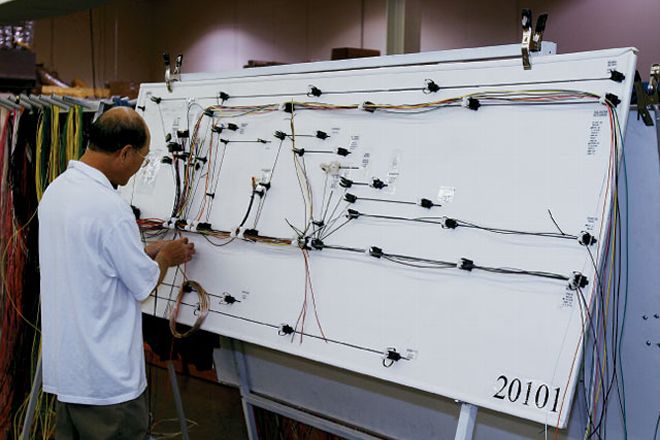}
\end{subfigure}
\caption{Two examples of a cable harnessing operation.}
\label{fig:harnessing}
\end{figure}
Among the various tasks typically carried out in the shop-floor, the manipulation of flexible or articulated objects, e.g., cable harnessing operations, is particularly challenging \cite{HenrichWorn2000, SaadatNan2002, Smithetal2012, Jimenez2012}, as can be seen in Figure \ref{fig:harnessing}:
on the one hand, it is usually beneficial to accurately plan the expected cable configurations on the harnessing table \textit{in advance}, thus confirming requirement $R_3$;
on the other hand, it is often necessary to keep a cable firm using more than two grasping points and to re-route the wiring pattern, which -- when done collaboratively with a robot, for instance to place bundle retainers or junction fixtures -- leads to requirements $R_1$ and $R_2$ above.
     
In the literature, the problem of determining the 2D or 3D configuration of flexible or articulated objects has received much attention in the past few years \cite{Wakamatsuetal2006, Nairetal2017}, whereas the problem of obtaining a target configuration via manipulation has been explored in motion planning \cite{Yamakawaetal2013, Bodenhagen2014, Schulmanetal2016}.
However, in the context of HRC, perception and manipulation are only part of the challenges to address.
Conceptually speaking, the outcome of such approaches is a \textit{continuous mapping} in 2D or 3D space from an initial to a target object's configuration \cite{Milleretal2011, Berenson2013, Bodenhagen2014, Ahmadzadehetal2015}, subject to a number of simplifying hypotheses as far as object models are concerned \cite{HowardBekey1997, Mastrogiovannietal2004, Franketal2010, Elbrechter2011, Elbrechter2012}.
This observation leads to two further functional requirements. 
The robot must be able to: 
\begin{enumerate}
\item[$R_4$] represent object configurations adopting suitable modelling assumptions, and then segment the whole manipulation problem in simpler actions to \textit{appropriately} sequencing and monitoring, each action operating in-between two intermediate configurations;
\item[$R_5$] represent the actions to perform using a formalism allowing for plan executions that are robust with respect to unexpected events (e.g., the human operator suddenly intervenes), and modelling errors (e.g., not modelled objects to be removed from the workspace).
\end{enumerate}

In this paper, we consider articulated objects as suitable models for flexible objects \cite{Yamakawaetal2013}, and we address the following challenges:
(i) we provide two representation and planning models for the classification of articulated object configurations and the sequencing of  manipulation actions, using an OWL-DL ontology-based formalism and the Planning Domain Definition Language (PDDL) \cite{FoxLong2003}, and we test them using two state-of-the-art PDDL planners, namely Probe \cite{LipovetzkyGeffner2011} and Madagascar \cite{Rintanen2014}, as well as with the VAL plan validator \cite{Foxetal2005};
(ii) we embed such models in a reactive/deliberative architecture for HRC, referred to as \textsc{planHRC}, which takes human operator behaviours into account and is implemented on top of the ROSPlan \cite{Cashmoreetal2015} and MoveIt! \cite{moveit} frameworks; and
(iii) we discuss how perception assumptions and representation schemes impact on planning and execution in HRC scenarios.
The \textsc{planHRC} architecture has been deployed on a dual-arm Baxter manipulator, which is used in all of our experiments.

The paper is organised as follows.
Section \ref{sec:related_work} discusses relevant approaches for the work described here.
Section \ref{sec:problem_statement} introduces more formally the problem we address, as well as the scenario we consider. 
The \textsc{planHRC}'s architecture is described in detail in Section \ref{sec:architecture}, where the overall information flow, the representation and reasoning challenges, and the planning models are discussed.
Experiments to validate the architecture are described in Section \ref{sec:experimental_validation}.
Conclusions follow. 

\section{Background}
\label{sec:related_work}

\subsection{Planning Techniques in Human-Robot Cooperation}
\label{sec:planning_arch_soa}

A number of studies have been conducted to investigate the role and the acceptability of automated planning techniques in HRC scenarios.
As highlighted in a field study by Gombolay and colleagues, two factors are important to maximise human satisfaction in HRC \cite{Gombolayetal2013}:
on the one hand, humans must be allowed to choose their own tasks freely, i.e., without them being assigned by an algorithm, subject to the fact that the cooperation is successful;
on the other hand, the overall system's (i.e., the human-robot team's) performance must be at high standards.   
It is noteworthy that these two factors may conflict in case of a \textit{lazy} or \textit{not focused} human attitude.  
However, when required to trade-off between them, humans show a strong preference for system's performance over their own freedom.
This study well fits with the requirements $R_1$, $R_2$ and $R_3$ outlined above, and opens up to an idea of a collaborative robot as a device not only able to \textit{aid} human workers, but also capable of \textit{keeping them in focus} and steering the cooperation towards its objectives if deviations occur. 

As a follow-up of the work discussed in \cite{Gombolayetal2013}, a study about the actual amount of control a human worker would like to have when collaborating with a robot has been reported in \cite{Gombolayetal2014}. 
The main finding of this study is that human workers tend not to prefer a total control of the cooperation process, rather they opt for partial control.
This is confirmed by the fact that the overall team's performance seems higher when the robot determines what actions must be carried out by the human.
As a consequence, a key factor for the acceptance of collaborative robots is finding a sensible -- yet efficient -- trade-off between performance and human control. 

In order to determine such trade-off, which may depend on the peculiar emotional or physical status of the human worker, one possibility is to encode in the planning process her/his preferences as far as tasks and operations are concerned \cite{Gombolayetal2015}.
In a first series of experiments, the use of human preferences in the planning algorithm led to an overall \textit{decrease} in performance, correlated with human subjective perceptions of robots not in line with the main cooperation objectives.
This suggests that a subjective assessment of the HRC process tends to attribute major inefficiencies to robots, and confirms that this is a crucial aspect for the applicability of collaborative robots in industrial scenarios.  
 
Techniques for HRC available in the literature target these issues only to a partial extent, positioning themselves at different levels in the trade-off scale outline above.
It is possible to identify two relevant trends for our work.

Approaches in the first class aim at defining cooperation models, i.e., data structures modelling the task to be jointly carried out, while keeping flexibility and human preferences into account \cite{Cirilloetal2010, Wilcoxetal2012, Tsarouchietal2016, JohannsmeierHaddadin2017, Ronconeetal2017, Sebastianietal2017, Darvishetal2017}.

A probabilistic planner is used in \cite{Cirilloetal2010} to sequence already defined partial plans, which include preferences about human preferred actions.
Once determined, the sequence of partial plans cannot be changed, therefore no flexibility for the human is allowed.
Such a limitation is partially overcome by the approach described in \cite{Wilcoxetal2012}, where an algorithm to adapt on-line both the action sequence and a number of action parameters is described.
This is achieved using a temporal formulation making use of preferences among actions, and using optimization techniques to identify the sequence best coping with preferences and constraints.
Still, the algorithm weighs more plan optimality (in terms of a reduced number of actions, or the time to complete the plan), and uses human preferences as soft constraints.
The approach by Tsarouchi and colleagues \cite{Tsarouchietal2016} assumes that a human and a robot co-worker operate in different workspaces.
The focus is on allocating tasks to the human or the robot depending on their preferences, suitability and availability, and the cooperation model is represented using an AND/OR graph.
Although human preferences are taken into account, task allocation is \textit{a priori} fixed and cannot be changed at run-time. 
A similar approach is considered in \cite{JohannsmeierHaddadin2017}, where the assumption about the separate workspaces is relaxed.
Hierarchical Task Models (HTMs) are used in \cite{Ronconeetal2017}, where the robot is given control on task allocation and execution is modelled using Partially Observable Markov Decision Processes (POMDPs).
However, the focus of this approach is on robot communication actions to enhance \textit{trust} in the human counterpart and to share a mutual understanding about the cooperation objectives. 
A similar approach is adopted in \cite{Sebastianietal2017}, where HTMs are substituted by Hierarchical Agent-based Task Planners (HATPs) and POMDPs are replaced by Petri Network Plans (PNPs).
However, differently from the approach in \cite{Ronconeetal2017}, the work by Sebastiani and colleagues support on-line changes during plan execution.
Finally, the work by Darvish and colleagues represents cooperation models using AND/OR graphs, and allows for a switch among different cooperation sequences at runtime \cite{Darvishetal2017}, therefore allowing humans to redefine the sequence of tasks among a predefined set of choices.
The human does not have to explicitly signal the switch to the robot, whereas the robot adapts to the new cooperation context reactively. 
 
The second class includes techniques focused on understanding, anticipating or learning human behaviours on-line \cite{Agostinietal2011, Koppulaetal2013, Karpasetal2015, LiuFisac2016, KwonSuh2014, CaccavaleFinzi2017}.
 
The work by Agostini and colleagues adopts classical planning approaches to determine an appropriate sequence of actions, given a model of the cooperation defined as a domain and a specific problem to solve \cite{Agostinietal2011}.
At runtime, the system ranks a predefined series of cause-effect events, e.g., observing their frequency as outcomes of human activities, and updates the cooperation model accordingly. 
Markov Decision Processes (MDPs) are used in \cite{Koppulaetal2013} to model the cooperation.
In particular, the human and the robot are part of a Markov decision game, and must cooperatively conclude the game, i.e., carrying out the cooperation process.
Human actions are detected on-line, which influences robot's behaviour at run-time. 
A similar approach, which takes into account temporal constraints among tasks, is discussed in \cite{Karpasetal2015}.
Statistical techniques to recognize human actions and to adapt an already available plan accordingly are presented in \cite{LiuFisac2016}.
Human deviations from the plan are detected.
When this happens, re-planning (including task allocation) occurs to achieve the cooperation objectives.
While the approaches discussed so far are quite conservative as far as robot's autonomy in the cooperation process is concerned, the work discussed in \cite{KwonSuh2014} exploits Bayesian networks to predict the occurrence and the timing of human actions.
Such a prediction is used to perform preparatory actions before an event even occurs.
While the overall system's performance is greatly improved, humans tend to be confused by the seemingly anticipative robot's behaviour.
Hierarchical Task Networks (HTNs) are used in \cite{CaccavaleFinzi2017} to embed communication actions in the cooperation process.
When specific deviations from the plan are detected, such communication actions enforce the adherence to the plan. 

\subsection{Rationale, Assumptions, Problem Statement, Reference Scenario}
\label{sec:problem_statement}

\textsc{planHRC} takes inspiration from the findings in \cite{Gombolayetal2013, Gombolayetal2014, Gombolayetal2015} to devise a cooperation model with the following characteristics:
\begin{itemize}
\item similarly to the work in \cite{Agostinietal2011}, the robot plans an appropriate, \textit{optimal}, sequence of actions to determine relevant intermediate configurations for an articulated object (considered as a simplified model for a flexible object like a cable), in order to determine a final target configuration, therefore coping with requirement $R_4$;
\item during plan execution, the robot always monitors the outcome of each action, and compares it with the target configuration to achieve, therefore limiting the burden on the human side \cite{Gombolayetal2014};
\item normally, the human can supervise robot actions: when a robot action is not successful, or a plan cannot be found, the human can intervene on the robot's behalf performing her/his preferred action sequence \cite{Gombolayetal2015}, therefore meeting $R_1$ and $R_2$;
\item at any time, the human can intervene (e.g., performing an action the robot was tasked with, or changing the articulated object's configuration), and the robot adapts to the new situation, in accordance with \cite{Gombolayetal2015} and requirements $R_3$ and $R_5$. 
\end{itemize}

More specifically, the problem we consider in this paper is three-fold:
(i) given a target object's configuration, determining a plan as an ordered set of actions:
\begin{equation}
a = \{a_1, \ldots, a_i, \ldots, a_N; \prec\},
\label{eq:plan}
\end{equation}
where each action $a_i$ involves one or more manipulation operations to be executed by a dual-arm robot,
(ii) designing a planning and execution architecture for the manipulation of articulated objects, which is \textit{efficient} and \textit{flexible} in terms of perceptual features, their representation and action planning, and
(iii) seamlessly integrating human actions \textit{in the loop}, allowing the robot to adapt to novel, not planned beforehand, object's configurations online.

In order to provide \textsc{planHRC} with such a features, we pose a number of assumptions described as follows:
\begin{itemize}
\item[$A_1$] flexible objects are modelled as articulated objects with a number of links and joints, as it is customary also for computational reasons \cite{Yamakawaetal2013}; we assume an inertial behaviour, i.e., rotating one link causes the movement of all upstream or downstream links, depending on the rotation joint; while this assumptions may not hold for soft flexible objects, it may reasonably hold for a wide range of bendable cables or objects;
\item[$A_2$] the effects of gravity on the articulated object's configurations are not considered, and the object is located on a table during all operations; 
\item[$A_3$] we do not assume in this paper any specific grasping or manipulation strategies to obtain a target object's configuration starting from another configuration; however, we do consider when an action $a_i$ cannot be completed because of unexpected events or modelling omissions;
\item[$A_4$] perception of articulated objects is affected by noise, but the \textit{symbol grounding problem}, i.e., the association between perceptual features and the corresponding symbols in the robot's knowledge representation system \cite{Harnad1990}, is assumed to be known.
\end{itemize}

On the basis of assumption $A_1$, we focus on articulated objects only. 
As anticipated above, we need to represent object's configurations.
We define an articulated object as a $2$-ple $\alpha = \left(\mathcal{L}, \mathcal{J}\right)$, where $\mathcal{L}$ is the ordered set of its $|L|$ links, i.e., $\mathcal{L} = \{l_1, \ldots, l_j, \ldots, l_{|L|}; \prec\}$, and $\mathcal{J}$ is the ordered set of its $|J|$ joints, i.e., $\mathcal{J} = \{j_1, \ldots, j_j, \ldots, j_{|J|}; \prec\}$.  
Each link $l_j \in \mathcal{L}$ is characterized by two parameters, namely a length $\lambda_l$ and an orientation $\theta_l$. 
We allow only for a limited number of possible orientation values.
This induces an ordered set $O$ of $|O|$ allowed orientation values, i.e., $O = \{o_1, \ldots, o_{|O|}; \prec\}$, such that an orientation $\theta_l$ can assume values in $O$.
We observe that in a cable harnessing operation this is only a minor simplification, since operators tend to distribute cables along predefined directions. 
Given a link $l_j$, we define two sets, namely $up(l_j)$ and $down(l_j)$, such that the former is made of upstream links, i.e., from $l_1$ to $l_{j-1}$, whereas the latter includes downstream links from $l_{j+1}$ to $l_{|J|}$. 

Orientations can be expressed with respect to an \textit{absolute}, possibly robot-centred reference frame, or -- less intuitively -- \textit{relative} to each other, for instance $\theta_{l_i}$ can represent the rotation with respect to $\theta_{l_{i-1}}$.
At a first glance, the absolute representation seems preferable because it leads to the direct perception of links and their orientations with respect to a robot-centred reference frame, whereas the set of absolute orientations constitute the overall object's configuration.
When a sequence of manipulation actions are planned, changing one absolute orientation requires -- in principle -- the \textit{propagation} of such change upstream or downstream the object via joint connections, which (hypothesis $H_1$) is expected to increase the computational burden on the reasoner and ($H_2$) may lead to suboptimal or redundant action sequences, because the propagation may jeopardise the effects of previous actions in the plan, or to sequences which cannot be fully understood by the human.
On the contrary, the less intuitive \textit{relative} approach assumes the direct perception of the relative orientations between pairwise links, and thus the overall object's configuration is made up of \textit{incremental} rotations. 
In this case, ($H_3$) computation is expected to be less demanding, since there is no need to propagate one change in orientation to upstream or downstream links, and therefore ($H_4$) actions on different links tend to be planned sequentially.
This has obvious advantages since it leads to shorter plans (on average), which could be further shortened by combining together action sub-sequences (e.g., two subsequent reorientations of $45$ $deg$ consolidated as one $90$ $deg$ single action), and to easy-to-understand plans.

\begin{figure}[t!]
\centering
\includegraphics[width=80mm]{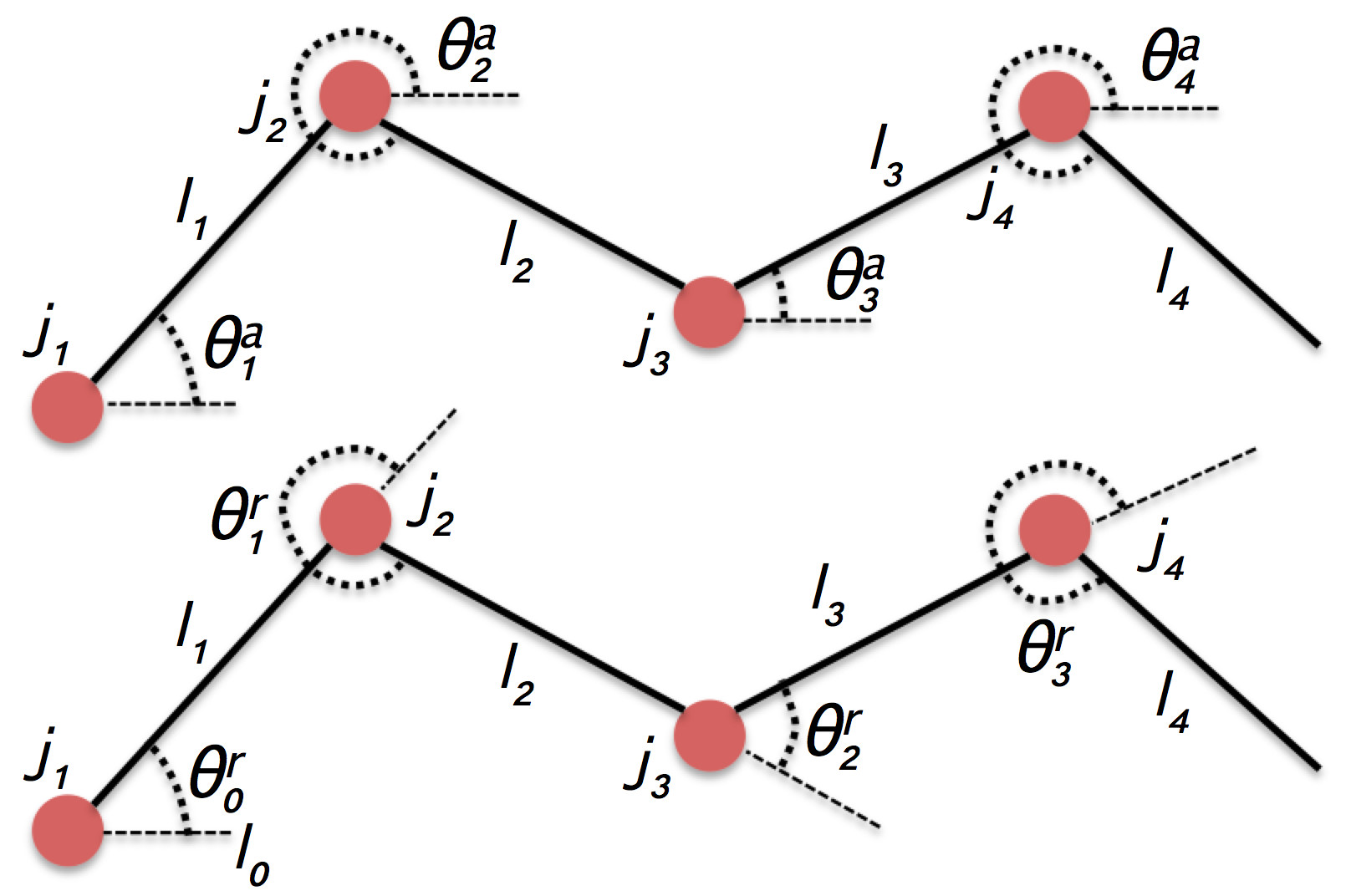}
\caption{Two possible representations: absolute (top) and relative (bottom).}
\label{fig:representation}
\end{figure}
If the object is represented using absolute orientations (Figure \ref{fig:representation} on the top), then its configuration is a $|L|$-ple:
\begin{equation}
\mathcal{C}_{\alpha, absolute} = \left(\theta^a_1, \ldots, \theta^a_l, \ldots, \theta^a_{|L|}\right), 
\label{eq:absolute}
\end{equation}
where it is intended that the generic element $\theta^a_l$ is expressed with respect to an absolute reference frame. 
Otherwise, if relative angles are used (Figure \ref{fig:representation} on the bottom), then the configuration must be \textit{augmented} with an initial \textit{virtual} link $l_0$ in order to define a reference frame, and therefore:
\begin{equation}
\mathcal{C}_{\alpha, relative} = \left(\theta^r_{0,virtual}, \theta^r_1, \ldots, \theta^r_l, \ldots, \theta^r_{|L|} \right).
\label{eq:relative}
\end{equation}
In principle, while the relative representation could model an object's configuration with one joint less compared to the absolute representation, the resulting configuration would not be unique (indeed there were infinitely many), since the object would maintain pairwise relative orientations between its links even when rotated \textit{as a whole}.
Therefore, an initial virtual reference link is added to the chain.

\begin{figure}[t!]
\centering
\includegraphics[width=80mm]{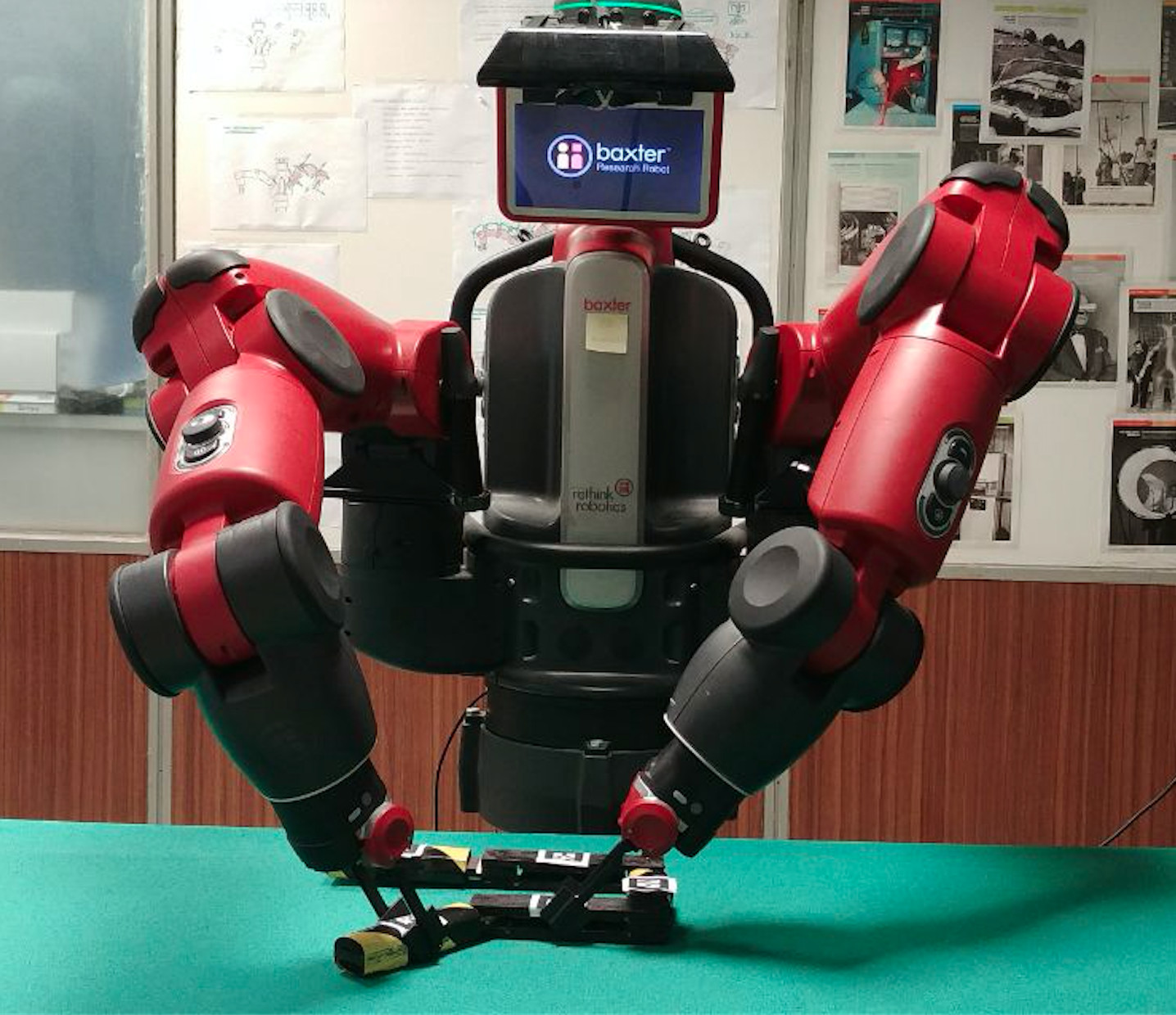}
\caption{The experimental scenario: a Baxter dual-arm manipulator operating on an articulated object.}
\label{fig:scenario}
\end{figure}
In order to comply with assumption $A_2$, we set up an experimental scenario where a Baxter dual-arm manipulator operates on articulated objects located on a table in front of it (Figure \ref{fig:scenario}).
Rotation operations occur only around axes centred on the object's joints and perpendicular to the table where the object is located.
We have crafted a wooden articulated object made up of $|L| = 5$ $15.5$ $cm$ long links, connected by $|J| = 4$ joints.
Links are $3$ $cm$ thick.
The object can be easily manipulated by the Baxter's standard grippers, which complies with assumption $A_3$.
To this aim, we adopt the MoveIt! framework. 
The Baxter is equipped with an RGB-D device located on top of its \textit{head} pointing downward to the table.
Only RGB information is used.
QR tags are fixed to each object's link, which is aimed at meeting assumption $A_4$.
Each QR code provides a 6D link pose, which directly maps to an absolute link orientation $\theta^a_l$.
Finally, if relative orientations are employed, we compute them by performing an algebraic sum between the two absolute poses of two consequent links, e.g., $\theta^r_1 = |\theta^a_2 - \theta^a_1|$.
A human can supervise robot operations and intervene when necessary from the other side of the table.

\section{\textsc{planHRC}'s Architecture}
\label{sec:architecture}

\subsection{Information Flow}
\label{sec:flow}

\begin{figure}[t!]
\centering
\includegraphics[width=100mm]{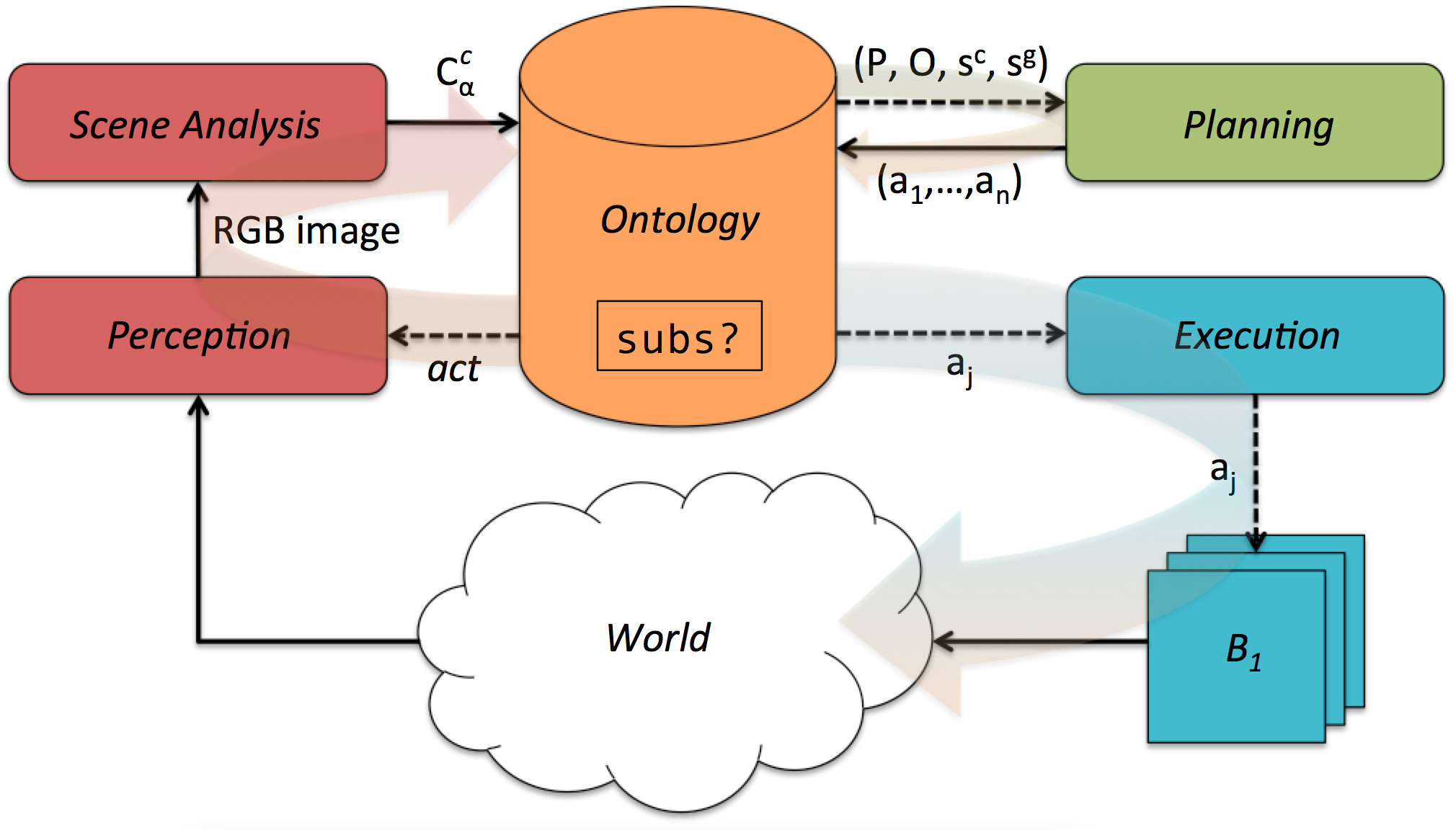}
\caption{The information flow in \textsc{planHRC}.}
\label{fig:architecture}
\end{figure}
\textsc{planHRC} allows for interleaved sensing, planning, execution and cooperation with a human using a number of parallel loops orchestrating the behaviour of different modules (Figure \ref{fig:architecture}).
Assuming that an articulated object $\alpha$ is located on the table in front of the robot, we want to modify its \textit{current} configuration $c^c_{\alpha}$ to obtain a goal configuration $c^g_{\alpha}$, which can be expressed as \eqref{eq:absolute} or \eqref{eq:relative}, i.e., it holds that $\mathcal{C}_{\alpha, absolute}(c^c_{\alpha})$ and $\mathcal{C}_{\alpha, absolute}(c^g_{\alpha})$, or $\mathcal{C}_{\alpha, relative}(c^c_{\alpha})$ and $\mathcal{C}_{\alpha, relative}(c^g_{\alpha})$, respectively.

The goal configuration $c^g_{\alpha}$ is encoded as assertional knowledge in an OWL-based \textit{Ontology} module \cite{Krotzsch2013}.
When this happens, the \textit{Perception} module is activated, and the Baxter's camera acquires an image of the workspace.
It is noteworthy that the \textit{Perception} module acquires images continuously, but for the sake of simplicity we treat each acquisition as if it were synchronous with action execution. 
The acquired image is filtered and registered, and appropriate artefact removal procedures are applied to produce an image suitable for feature extraction, which is fed to the \textit{Scene Analysis} module.  
A \textit{perceived} configuration $c^p_{\alpha}$ (i.e., the current configuration $c^c_{\alpha}$) is extracted from the image, and a representation of it stored in the \textit{Ontology} module.
Both $c^c_{\alpha}$ and $c^g_{\alpha}$ are represented using conjunctions of class instances, which model such predicates as $\mathsf{Connected}$, to indicate whether two links are connected by a joint, or $\mathsf{HasOrientation}$, to define angle orientations.  
If $c^c_{\alpha}$ and $c^g_{\alpha}$ \textit{are not compatible} then a planning process occurs.
In order to determine compatibility, we assume the presence of an operator $\mathcal{D}$ that, given a terminological or assertional element in the ontology, returns its description in OWL formalism.
Therefore, if the description of $c^c_{\alpha}$ is not \textit{subsumed} by the description of $c^g_{\alpha}$, i.e., it does not hold that $\mathcal{D}(c^c_{\alpha}) \sqsubseteq \mathcal{D}(c^g_{\alpha})$, the planner is invoked.
Specifically, a \textit{Planner} module is activated, which requires the definition of relevant predicates $\mathcal{P}_1, \ldots, \mathcal{P}_{|P|}$, and possible action types $\mathcal{A}_1, \ldots, \mathcal{A}_j, \ldots, \mathcal{A}_{|A|}$ in the form:
\begin{equation}
\mathcal{A}_j = \left(pre(\mathcal{A}_j), eff^-(\mathcal{A}_j), eff^+(\mathcal{A}_j)\right),
\label{eq:action}
\end{equation}
where $pre(\mathcal{A}_j)$ is the set of preconditions (in the form of predicates) for the action to be executable, $eff^-(\mathcal{A}_j)$ is the set of negative effects, i.e., predicates becoming false after action execution and $eff^+(\mathcal{A}_j)$ is the set of positive effects, i.e., predicates becoming true after execution.
For certain domains, it is useful to extend \eqref{eq:action} to allow for additional positive or negative effects, i.e., predicates becoming true or false in case certain additional conditions hold.
A conditional action can be modelled as:
\begin{equation}
\mathcal{A}_j = \left(pre(\mathcal{A}_j), eff^-(\mathcal{A}_j), eff^+(\mathcal{A}_j), pre_a(\mathcal{A}_j), eff^-_a(\mathcal{A}_j), eff^+_a(\mathcal{A}_j)\right),
\label{eq:condaction}
\end{equation} 
where $pre(\mathcal{A}_j)$, $eff^-(\mathcal{A}_j)$ and $eff^+(\mathcal{A}_j)$ are defined as before, $pre_a(\mathcal{A}_j)$ is the set of additional preconditions, whereas and $eff^-_a(\mathcal{A}_j)$ and $eff^+_a(\mathcal{A}_j)$ are the sets of additional effects subject to the validity of predicates in $pre_a(\mathcal{A}_j)$.
Furthermore, the \textit{Planner} requires a suitable description of the current state $s^c$ (including a description of $c^c_{\alpha}$) and the goal state $s^g$ (including $c^g_{\alpha}$), described using an appropriate set of ground predicates $p_1, \ldots, p_{|p|}$. 
This information, encoded partly in the terminological section and partly in the assertional section of the \textit{Ontology} module, is translated in a format the \textit{Planner} module can use, namely the Planning Domain Definition Language (PDDL) \cite{PDDL}.  

A plan, as formally described in \eqref{eq:plan}, is an ordered sequence of $N$ actions whose execution changes the current state from $s^c$ to $s^g$ through a set of intermediate states.
In a plan, each action may correspond to one or more scripted robot behaviours.
For example, rotating a link $l_{j+1}$ may require the robot to (i) keep the upstream link $l_j$ steady with its left gripper, and (ii) rotate $l_{j+1}$ of a certain amount with the right gripper.
Such sequence shall not be encoded in the planning process, thereby reducing planning cost, but demanded to an action execution module.  
If a plan is found, each action is encoded in the \textit{Ontology}, along with all the \textit{expected} intermediate states ${s^c = s^e_1}, s^e_2, \ldots, {s^g = s^e_{N+1}}$, which result from actions.
The \textit{Execution} module executes action by action activating the proper modules in the architecture, e.g., such \textit{behaviours} as motion planning, motion execution, obstacle avoidance or grasping. 

\begin{figure}[t!]
\centering
\includegraphics[width=120mm]{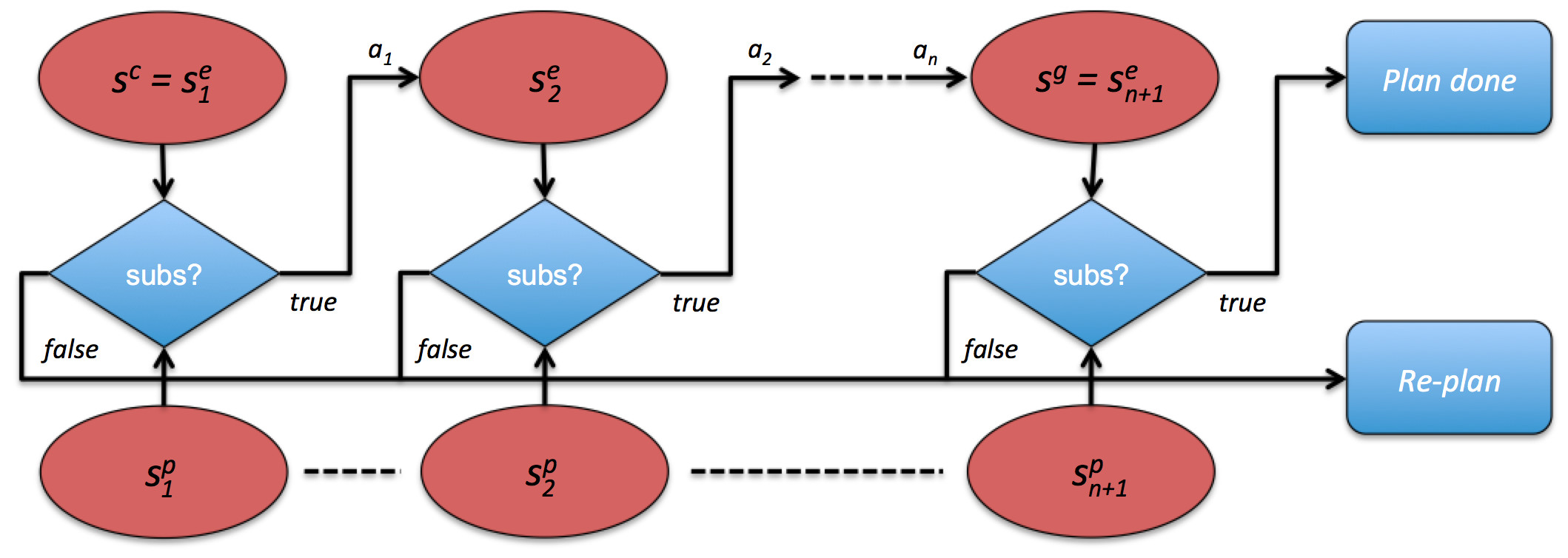}
\caption{The planning and execution pipeline.}
\label{fig:planning}
\end{figure}
Each action $a_j$ in a plan is assumed to transform a state $s^e_j$ into a state $s^e_{j+1}$, such that:
\begin{equation}
s^e_{j+1} = \left(s^e_j\setminus{eff}^-(a_j)\right) \cup {eff}^+(a_j).
\label{eq:statetransition}
\end{equation}
If $a_j$ has additional conditions, then \eqref{eq:statetransition} is modified as:
\begin{equation}
s^e_{j+1} = \left(s^e_j\setminus\left({eff}^-(a_j) \cup C^-(pre_a(a_j)) \right)\right) \cup \left({eff}^+(a_j) \cup C^+(pre_a(a_j)) \right), 
\label{eq:statetransitioncond}
\end{equation}
where conditions $C^-$ and $C^+$ return the sets ${eff}^-_a(a_j)$ and ${eff}^+_a(a_j)$, respectively, if the conditions in $pre_a(a_j)$ hold, and $\emptyset$ otherwise.  
Before the action is executed, the \textit{Ontology} module activates \textit{Perception} to acquire a new image.
Again, this induces a new perceived, current configuration $c^c_{\alpha}$. 
Every time this happens, two situations can happen:
if $c^c_{\alpha}$ corresponds to a current perceived state $s^c$ whose description is subsumed by the description of a state $s^e_{j-1}$ possibly generated applying an action $a_{j-1}$ or as a consequence of human intervention, i.e., $\mathcal{D}(s^c) \sqsubseteq \mathcal{D}(s^e_{j-1})$, then the execution continues with action $a_{j}$ until a state is reached which is subsumed by $\mathcal{D}(s^g)$;
otherwise, a new planning process occurs, considering the current state $s^c$ as a new initial state and keeping the previous goal state $s^g$.

A few remarks can be made.
When an action $a_j$ is executed, the expected intermediate state $s^e_{j}$ is treated as a set of \textit{normative} ground predicates, i.e., it defines the normal, expected state for $a_j$ to be feasible.
Whether $s^e_{j}$ is obtained as a result of a previous action, or with the help of the human operator is not relevant for $a_j$.
On the contrary, deviations from it are treated as violations and therefore the system tries to re-plan in order to reach a state compatible with $s^g$ starting from the current state.
As discussed above, violations can be of two kinds:
on the one hand, human interventions (i.e., object manipulations on robot's behalf) may lead to a current state $s^c$ not compatible with the expected intermediate state $s^e_{j}$, and therefore the robot should adapt by re-planning;
on the other hand, a robot may not be able to complete action $a_j$, e.g., due to a cluttered workspace \cite{Srivastavaetal2014} or the obstructing presence of the human operator \cite{HaddadinCroft2016}.
In the second case, if such an event were detected, the robot would re-plan starting from the current state, and possibly ask for the human operator's help to achieve a workable object's configuration.   
As a consequence, \textsc{planHRC} implements a policy according to which the overall system's performance is ensured by the use of state-of-the-art planning techniques, but it allows at any time the human operator to intervene and forces the robot to adapt its plan accordingly. 

Figure \ref{fig:planning} shows a graphical model of the information flow from the perspective of the planning process.

\subsection{Representation Models in the Ontology}
\label{sec:representation}

An ontology $\Sigma = \left(TBox, ABox \right)$ is a $2$-ple where the $TBox$ is a terminological taxonomy of axioms storing definitions of relevant classes and relationships within a domain, and the $ABox$ is an assertional taxonomy representing factual knowledge for such a domain.
Both $TBox$ and $ABox$ are described using the Description Logic formalism \cite{Baaderetal2003} through its computational variant referred to as the Web Ontology Language (OWL), and in particular OWL-DL \cite{OWL}, plus the addition of SWRL rules for deductive reasoning \cite{Horrocksetal2005}.

In \textsc{planHRC}, the ontology is used both \textit{off-line} and \textit{on-line} for different purposes\footnote{The OWL ontology used in this work is available for further evaluation and improvements at: \url{https://github.com/EmaroLab/OWL-ROSPlan/tree/master/rosplan\_knowledge\_base/}.}.
The off-line use is related to modelling the domain of articulated objects manipulation, in terms of \textit{types}, \textit{predicates}, \textit{operators}, \textit{states}, \textit{problems} and \textit{plans}.
The on-line use serves two purposes:
on the one hand, to represent relevant object's configurations, such as the current $c^c_{\alpha}$ and the goal $c^g_{\alpha}$ configurations, as well as specific actions to perform using classes and relationships defined in the $TBox$;
on the other hand, to apply such reasoning techniques as \textit{instance checking} to the representation, e.g., to determine whether an action $a_j$ assumes an expected planning state $s^e_{j}$ which is compatible with the perceived current state $s^c$, as described in Figure \ref{fig:planning}.

To this aim, the taxonomy in the $TBox$ models \textit{types}, which are used by the \textit{Planner} module when processing PDDL syntax, as primitive classes derived from $\mathsf{Type} \sqsubseteq \top$, such as $\mathsf{Link} \sqsubseteq\mathsf{Type}$ and $\mathsf{Joint} \sqsubseteq\mathsf{Type}$, as well as $\mathsf{Orientation} \sqsubseteq\mathsf{Integer}$ (represented using \textit{degrees}).
Relevant \textit{predicates} are modelled as classes derived from $\mathsf{Predicate} \sqsubseteq \top$.
For instance, $\mathsf{Connected}$ is used to relate a $\mathsf{Joint}$ to a $\mathsf{Link}$, as:
\begin{equation}
\begin{split}
\mathsf{Connected} \sqsubseteq & \, \mathsf{Predicate} \, \sqcap\\
& \, \exists \mathsf{arg1}.\mathsf{Joint} \, \sqcap \, \mathsf{=_1} \mathsf{arg1} \, \sqcap\\
& \, \exists \mathsf{arg2}.\mathsf{Link} \, \sqcap \, \mathsf{=_1} \mathsf{arg2}.
\end{split}
\label{eq:connected}
\end{equation}
In the description, $\mathsf{Connected}$ has two arguments, namely $\mathsf{arg1}$ and $\mathsf{arg2}$, which relate exactly \textit{one} $\mathsf{Joint}$ with \textit{one} $\mathsf{Link}$.
While this perfectly fits with the first and last link composing the articulated object, intermediate links will be modelled using two $\mathsf{Connected}$ predicates, for the downstream and upstream links, respectively.   
In order to specify the orientation associated with a $\mathsf{Link}$ with respect to a $\mathsf{Joint}$:
\begin{equation}
\begin{split}
\mathsf{HasOrientation} \sqsubseteq & \, \mathsf{Predicate} \, \sqcap\\
& \, \exists \mathsf{arg1}.\mathsf{Joint} \, \sqcap \, \mathsf{=_1} \mathsf{arg1} \, \sqcap\\
& \, \exists \mathsf{arg2}.\mathsf{Orientation} \, \sqcap \, \mathsf{=_1} \mathsf{arg2},
\end{split}
\label{eq:hasorientation}
\end{equation}
where the semantics associated with $\mathsf{arg2}$ depends on whether we adopt absolute or relative angles, as described in Section \ref{sec:problem_statement}.
As discussed above, in the planning process the value an orientation can take is related to the set $\mathcal{O}$ of allowed orientation values.
This is done to reduce the state space involved in the planning process, and the sensitivity of \textsc{planHRC}'s computational performance with respect to the cardinality of the set $\mathcal{O}$ is analysed as part of experimental validation.
Irrespectively whether orientations are absolute or relative, $\mathcal{O}$ is represented as a collection of predicates relating pairwise values:
\begin{equation}
\begin{split}
\mathsf{OrientationOrd} \sqsubseteq & \, \mathsf{Predicate} \, \sqcap\\
& \, \exists \mathsf{arg1}.\mathsf{Orientation} \, \sqcap \, \mathsf{=_1} \mathsf{arg1} \, \sqcap\\
& \, \exists \mathsf{arg2}.\mathsf{Orientation} \, \sqcap \, \mathsf{=_1} \mathsf{arg2}.
\end{split}
\label{eq:orientationord}
\end{equation}
For instance, if only two possible orientations are allowed, namely $30$ $deg$ and $45$ $deg$, $\mathcal{O}$ can be modelled using only \textit{one} predicate $\mathsf{OrientationOrd(ord\_30\_45)}$ such that:
\begin{equation}
\begin{split}
&\mathsf{arg1(ord\_30\_45, 30)},\\
&\mathsf{arg2(ord\_30\_45, 45)},
\end{split}
\label{eq:orientationord_example}
\end{equation}
where it is intended that the orientations $30$ $deg$ and $45$ $deg$ are associated with $\mathsf{arg1}$ and $\mathsf{arg2}$, respectively. 
Other predicates are described in a similar way.
As it will become evident in Section \ref{sec:planning}, the absolute representation of orientations requires the use of \textit{conditional operators} in PDDL.
These are modelled in the $TBox$ using \textit{conditional predicates} to be mapped to PDDL operators:
\begin{equation}
\begin{split}
\mathsf{CondPredicate} \sqsubseteq & \, \mathsf{Predicate} \, \sqcap\\
& \, \exists \mathsf{forall}.\mathsf{Type} \, \sqcap \, \mathsf{\geq_1} \mathsf{forall} \, \sqcap\\  
& \, \exists \mathsf{when}.\mathsf{Predicate} \, \sqcap \, \mathsf{\geq_1} \mathsf{when} \, \sqcap\\  
& \, \exists \mathsf{{eff-}\_a}.\mathsf{Predicate} \, \sqcap \, \mathsf{\geq_1} \mathsf{{eff-}\_a}, \, \sqcap\\ 
& \, \exists \mathsf{{eff+}\_a}.\mathsf{Predicate} \, \sqcap \, \mathsf{\geq_1} \mathsf{{eff+}\_a}, 
\end{split}
\label{eq:condpredicate}
\end{equation}
where the intuitive meaning is that for all $\mathsf{Type}$ individuals specified in the relationship, when specific $\mathsf{Predicate}$ individuals hold, the additional effects must be considered. 

It is possible to define $\mathsf{Action} \sqsubseteq \top$ as:
\begin{equation}
\begin{split}
\mathsf{Action} \sqsubseteq & \, \top \, \sqcap\\
& \, \exists \mathsf{params}.\mathsf{Type} \, \sqcap \, \mathsf{\geq_1} \mathsf{params} \, \sqcap\\
& \, \exists \mathsf{pre}.\mathsf{Predicate} \, \sqcap \, \mathsf{\geq_1} \mathsf{pre} \, \sqcap\\  
& \, \exists \mathsf{eff-}.\mathsf{Predicate} \, \sqcap \, \mathsf{\geq_1} \mathsf{eff-} \, \sqcap\\  
& \, \exists \mathsf{eff+}.\mathsf{Predicate} \, \sqcap \, \mathsf{\geq_1} \mathsf{eff+} \, \sqcap\\
& \, \exists \mathsf{condEff}.\mathsf{CondPredicate}.  
\label{eq:dl_action}
\end{split}
\end{equation}
In \eqref{eq:dl_action}, we do not assume the presence of a relationship $\mathsf{condEff}$ to the aim of modelling both actions and conditional actions using the same definition.
In our $TBox$, two actions are defined, namely $\mathsf{RotateClockwise} \sqsubseteq \mathsf{Action}$, which increases a link orientation of a certain amount, and $\mathsf{RotateAntiClockwise} \sqsubseteq \mathsf{Action}$, which does the opposite.

One notable predicate used as part of conditional effects is $\mathsf{Affected}$, which models how changing a link orientation propagates via connected upstream or downstream joints:
\begin{equation}
\begin{split}
\mathsf{Affected} \sqsubseteq & \, \mathsf{Predicate} \, \sqcap\\
& \, \exists \mathsf{arg1}.\mathsf{Joint} \, \sqcap \, \mathsf{=_1} \mathsf{arg1} \, \sqcap\\
& \, \exists \mathsf{arg2}.\mathsf{Link} \, \sqcap \, \mathsf{=_1} \mathsf{arg2} \, \sqcap\\
& \, \exists \mathsf{arg3}.\mathsf{Joint} \, \sqcap \, \mathsf{=_1} \mathsf{arg3},
\end{split}
\label{eq:affected}
\end{equation}  
which states that a change in orientation related to the joint in $\mathsf{arg1}$ is affected by rotations of joint specified in $\mathsf{arg3}$, as obtained when operating on the link in $\mathsf{arg2}$.

Likewise, any \textit{state} $s$ (perceived, current, predicted or expected) is represented as a set of predicates:
\begin{equation}
\begin{split}
\mathsf{State} \sqsubseteq & \, \top \, \sqcap\\
& \, \exists \mathsf{madeof}.\mathsf{Predicate} \, \sqcap \, \mathsf{\geq_1} \mathsf{madeof},  
\end{split}
\label{eq:state}
\end{equation}
through the relationship $\mathsf{madeof}$, which must include at least one $\mathsf{Predicate}$ for the state to be formally expressed. 
As a consequence, a planning \textit{problem} is modelled as having an initial and a goal $\mathsf{State}$:
\begin{equation}
\begin{split}
\mathsf{Problem} \sqsubseteq & \, \top \, \sqcap\\
& \, \exists \mathsf{init}.\mathsf{State} \, \sqcap \, \mathsf{=_1} \mathsf{init} \, \sqcap\\  
& \, \exists \mathsf{goal}.\mathsf{State} \, \sqcap \, \mathsf{=_1} \mathsf{goal}.
\end{split}
\label{eq:problem}
\end{equation}
Finally, a $\mathsf{Plan} \sqsubseteq \top$ is made up of actions: 
\begin{equation}
\begin{split}
\mathsf{Plan} \sqsubseteq & \, \top \, \sqcap\\
& \, \exists \mathsf{madeof}.\mathsf{Action} \, \sqcap \, \mathsf{\geq_1} \mathsf{madeof}.  
\end{split}
\label{eq:dl_plan}
\end{equation}

On-line, the $ABox$ is updated each time a new image is acquired by the \textit{Perception} module, and maintains descriptions in the form of assertions possibly classified as being instances of the classes defined in the $TBox$ terminology.
Let us describe what happens at each iteration with an example.
If the robot perceived an object configuration like the one in Figure \ref{fig:representation} on the top, four $\mathsf{Link}$ instances:
\begin{equation}
\begin{split}
\mathsf{Link(l1)}\quad
\mathsf{Link(l2)}\quad
\mathsf{Link(l3)}\quad
\mathsf{Link(l4)}
\end{split}
\end{equation}
and four $\mathsf{Joint}$ instances:
\begin{equation}
\begin{split}
\mathsf{Joint(j1)}\quad
\mathsf{Joint(j2)}\quad
\mathsf{Joint(j3)}\quad
\mathsf{Joint(j4)}
\end{split}
\end{equation}
would be necessary to represent it.
Then, the object's structure would be modelled as a description including the set of predicate instances:
\begin{equation}
\begin{split}
&\mathsf{Connected(connected\_j1\_l1)}\\
&\mathsf{Connected(connected\_j2\_l1)}\\
&\mathsf{Connected(connected\_j2\_l2)}\\
&\ldots\\
&\mathsf{Connected(connected\_j4\_l4)}.
\end{split}
\end{equation}
where $\mathsf{connected\_j1\_l1}$ is such that:
\begin{equation}
\begin{split}
&\mathsf{arg1(connected\_j1\_l1, j1)}\\
&\mathsf{arg2(connected\_j1\_l1, l1)},
\end{split}
\end{equation} 
as specified in \eqref{eq:connected}. 
Other \textit{Predicate} instances can be generated in a similar way.
Furthermore, assuming that $\theta^a_1 = 45$ $deg$, $\theta^a_2 = {330}$ $deg$, $\theta^a_3 = 30$ $deg$ and $\theta^a_4 = {315}$ $deg$, orientations would be represented as:
\begin{equation}
\begin{split}
&\mathsf{HasOrientation(has\_orientation\_j1\_45)}\\
&\mathsf{HasOrientation(has\_orientation\_j2\_{330})}\\
&\mathsf{HasOrientation(has\_orientation\_j3\_30)}\\
&\mathsf{HasOrientation(has\_orientation\_j4\_{315})},
\end{split}
\end{equation}
where, focusing on $\mathsf{arg2}$ only:
\begin{equation}
\begin{split}
&\mathsf{arg2(has\_orientation\_j1\_45, 45)}\\
&\mathsf{arg2(has\_orientation\_j2\_{330}, {330})}\\
&\mathsf{arg2(has\_orientation\_j3\_30, 30)}\\
&\mathsf{arg2(has\_orientation\_j4\_{315}, {315})},
\end{split}
\end{equation}
All such $\mathsf{Connected}$ and $\mathsf{HasOrientation}$ instances would contribute to the definition of the current state $\mathsf{State(state\_c)}$ by means of a set of assertions like:
\begin{equation}
\begin{split}
&\mathsf{madeof(state\_c, connected\_j1\_l1)}\\
&\mathsf{madeof(state\_c, connected\_j2\_l1)}\\
&\ldots\\
&\mathsf{madeof(state\_c, has\_orientation\_j4\_{315}, {315})},
\end{split}
\end{equation}
as foreseen by \eqref{eq:state}.
Similar descriptions for problems and, after the planning process occurs, for plans, can be introduced as well.

When a new goal state $\mathsf{State(state\_g)}$ is encoded in the ontology, a new $\mathsf{Problem(problem\_c)}$ is created, such that, according to \eqref{eq:problem}:
\begin{equation}
\begin{split}
&\mathsf{init(problem\_c, state\_c)}\\
&\mathsf{goal(problem\_c, state\_g)},
\end{split}
\label{eq:problem_instance}
\end{equation}
and the \textit{Planner} module is activated. 
A translation process occurs, which generates the proper PDDL formulation by querying the $TBox$ (to generate the PDDL \textit{domain}) and the $ABox$ (to generate the PDDL \textit{problem}).
Each class in the $TBox$ roughly corresponds to a section of the domain, whereas $\mathsf{state\_c}$ and $\mathsf{state\_g}$ in the $ABox$ define the initialisation and goal sections of a problem.

After a plan has been found and validated (see Section \ref{sec:planning}), each action is encoded back in the ontology as an instance of $\mathsf{Action}$, and therefore all relationships $\mathsf{param}$, $\mathsf{pre}$, $\mathsf{eff-}$ and $\mathsf{eff+}$ are specified in terms of $\mathsf{Type}$ and $\mathsf{Predicate}$ instances.
If an action has conditional effects, also $\mathsf{condEff}$ is determined.  
As a consequence, a set of intermediate expected states is create as:
\begin{equation}
\begin{split}
&\mathsf{State(state\_e\_1)}\\
&\mathsf{State(state\_e\_2)}\\
&\ldots\\
&\mathsf{State(state\_e\_{{n+1}})}
\end{split}
\end{equation}
as described in Section \ref{sec:flow}.
In particular, $\mathsf{state\_e\_1} \equiv \mathsf{state\_c}$, $\mathsf{state\_e\_{{n+1}}} \equiv \mathsf{state\_g}$, and the intermediate expected states are generated using \eqref{eq:statetransition} and \eqref{eq:statetransitioncond}.
When $\mathsf{State}$ individuals are generated, the \textit{Execution} module is activated.
   
\subsection{Reasoning in the Ontology, or the Cooperation Model}
\label{sec:reasoning}

As anticipated in Section \ref{sec:problem_statement}, and in accordance with the findings in \cite{Gombolayetal2013, Gombolayetal2014, Gombolayetal2015}, the human-robot cooperation model implemented in \textsc{planHRC} foresees that:
(i) the robot determines a plan maximizing some performance indicator in terms of number of actions and/or time-to-completion;
(ii) the robot monitors the execution of each action in the plan;
(iii) during normal work flow, the human supervises robot actions; and
(iv) the human can intervene to cope with robot's failures in action planning or execution, or to perform tasks asynchronously and in parallel to robot activities.   
The model unfolding is based on monitoring the state transitions in \eqref{eq:statetransition} and \eqref{eq:statetransitioncond} and their failures.
Independently of the presence of conditional effects in an action $a_j$, two cases are possible after the action is submitted to the \textit{Execution} module: it cannot be executed (or it is executed only in part) or it is carried out successfully.

The first case originates from motion planning or execution issues, e.g., because of a cluttered workspace \cite{Srivastavaetal2014} or to prevent any harm to the human operator \cite{Deneietal2015, HaddadinCroft2016, Darvishetal2017}.
If motion issues occur, \textsc{planHRC} does not generate a state compatible with $s_{j+1}^e$.
However, this does not necessarily mean that the current state $s^c$ is still compatible with the previous assessed state $s_{j}^e$, i.e., $\mathcal{D}(\mathsf{state\_c}) \sqsubseteq \mathcal{D}(\mathsf{state\_e\_j})$ may not hold, because the robot may have completed only part of the action.
In this case, a new current state $s^c$ is acquired.
If there is any intermediate expected state $s_i^e$ comparable with $s^c$, i.e., there exists in the ontology an individual $\mathsf{State(state\_e\_i)}$ such that $\mathcal{D}(\mathsf{state\_c}) \sqsubseteq \mathcal{D}(\mathsf{state\_e\_i})$, then execution resumes from action $a_{i+1}$; otherwise, it is necessary to invoke again the \textit{Planner} module using $\mathsf{state\_c}$ and $\mathsf{state\_g}$, and obtain a new plan.     

In the second case, action $a_j$ is considered to be successful from the point of view of motion execution.
Still, the outcome may or may not be compatible with the expected state $s_{j+1}^e$, e.g., due to not modelled effects.
This state is observable as the current state $s^c$.
However, although $\mathcal{D}(\mathsf{state\_c}) \sqsubseteq \mathcal{D}(\mathsf{state\_e\_{j+1}})$ does not hold, it may happen that $s^c$ could be appropriate for the next action $a_{j+1}$ to occur. 
In particular, for $a_{j+1}$ to be executable, it must hold that $s^c \sqsubseteq pre(a_{j+1})$, i.e., $\mathcal{D}(\mathsf{state\_c}) \sqsubseteq \mathcal{D}(\mathsf{action\_{j+1}.pre})$.
We treat the set of predicates in $pre(a_{j+1})$ as normative conditions for $a_{j+1}$, regardless whether the expected state $s_{j+1}^e$ is generated as the outcome of the previous action $a_{j}$.
If $s^c \sqsubseteq pre(a_{j+1})$ does not hold, we must check whether there is any intermediate expected state $s_i^e$ comparable with $s^c$: if it is the case, execution resumes from action $a_{i+1}$; otherwise, re-planning is necessary.

In summary, human intervention is strictly necessary when a plan cannot be found.
However, any human action is implicitly considered every time the current state does not comply with normative predicates. 

\subsection{Planning Models}
\label{sec:planning}

As anticipated in Section \ref{sec:problem_statement}, orientations can be expressed using an absolute or relative reference frame, meaning that in the first case each link's orientation is expressed with respect to an externally-defined reference frame, whereas in the second case it is expressed with respect to another link's orientation, e.g., the upstream one.
These two possibilities lead to two planning models, with different semantics, which are characterized by different properties as far as (i) obtained plan, (ii) computational load of the planning process, and (iii) ease of execution for the robot, are concerned.

For the sake of simplicity, we present the relative formulation first, and then the absolute one.  
The relative formulation employs the \texttt{:STRIPS} subset of PDDL, extended with \texttt{:equalities} and \texttt{:negative-preconditions}, whereas the absolute version requires also the use of \texttt{:conditional-effects}.
Notably, the problem we are interested in assumes a sort of granularity discretization of angular orientations, hence there is no practical necessity for continuous or hybrid planning models \cite{FoxLong2006}.
Therefore, PDDL constitutes an appropriate level of abstraction\footnote{Examples of planning domains and problems can be found at \url{https://github.com/EmaroLab/paco_actions}.}. 

As discussed when introducing assumption $A_1$, our model assumes a sort of inertial behaviour, i.e., rotating one link affects the orientation of upstream or downstream links as well.
In fact, given a link $l_j$ to rotate (clockwise or anticlockwise), two rotation actions are possible:
on the one hand, if link $l_{j-1}$ is kept still and $l_j$ is rotated (clockwise or anticlockwise), then all links in $down(l_j)$ rotate (clockwise or anticlockwise) and are displaced as well; on the other hand, if link $l_{j+1}$ is kept still, all links in $up(l_j)$ are rotated (clockwise or anticlockwise) and displaced.

From a planning perspective, each rotation action (either clockwise or anticlockwise) changing an angle $\theta^r_j$ referring to a relative orientation does not affect any other orientations of links in $up(l_j)$ or $down(l_j)$, since all of them are relative to each other, and therefore the planning process is computationally less demanding. 
However, since actions are expected to be based on link orientations grounded with respect to a robot-centred reference frame, i.e., absolute in terms of pairwise link orientations, a conversion must be performed, which may be greatly affected by perceptual noise, therefore leading to inaccurate or even inconsistent representations.
In the absolute formulation, $\theta^a_j$ is considered absolute, and therefore it can be associated directly with robot actions. 
Unfortunately, this means that each action changing $\theta^a_j$ does affect numerically all other orientations of links in $up(l_j)$ or $down(l_j)$ in the representation, which must be kept track of using conditional effects in the planning domain.
It is noteworthy that the use of advanced PDDL features, such as conditional effects, may allow for a more accurate representation of the domain but, at the same time, it may reduce the number of planners able to reason on the model, and their efficiency.
 
\begin{figure}[t!]
\begin{mdframed}
\footnotesize
\texttt{
\begin{tabbing}
(:\=ac\=ti\=on\:\:\:RotateClockwise\\
\> :parameters (?l1 ?l2 - Link\\
\>\> ?j1 - Joint ?o1 ?o2 - Orientation)\\
\> :precondition (and \\
\>\> (Connected ?j1 ?l1)\\
\>\> (Connected ?j1 ?l2)\\
\>\> (not (= ?l1 ?l2))\\
\>\> (HasOrientation ?o1 ?j1)\\
\>\> (OrientationOrd ?o1 ?o2))\\
\> :effect (and \\
\>\> (not (HasOrientation ?o1 ?j1)) \\
\>\> (HasOrientation ?o2 ?j1))\\
)
\end{tabbing}
}
\end{mdframed}
\caption{The \textit{relative} version of $\mathsf{RotateClockwise}$ in PDDL.}
\label{fig:rotate_clockwise_basic}
\end{figure}
\textit{Relative formulation}.
As described in Section \ref{sec:problem_statement}, an articulated object $\alpha$ is represented using two ordered sets of links and joints.
We use a \texttt{Connected} predicate modelled as described in \eqref{eq:connected} to describe the sequence of links in terms of binary relationships each one involving a link $l_j$ and a joint $j_{j+1}$, which induces a pairwise connection between two links, namely $l_j$ and $l_{j+1}$, since they share the same joint $j_{j+1}$.
The orientation of a link $l_j$ is associated with the corresponding joint $j_j$ and corresponds to an angle $\theta^r_j$, which ranges between $0$ and $359$ $deg$, using the predicate \texttt{HasOrientation} as specified in \eqref{eq:hasorientation}.
As anticipated above, this formulation assumes that link orientations are expressed incrementally relative to each other.
This means that the robot's perception system is expected to provide the \textit{Ontology} module with the set of relative link orientations as primitive information.  
If absolute link orientations are not available, the object's configuration $\mathcal{C}_{\alpha, absolute}$ can be computed applying forward kinematics formulas using relative orientations and link lengths.
If noise affects the perception of link orientations, as it typically does, the reconstruction of the object's configuration may differ from the real one, and this worsens with link lengths.
However, this model significantly simplifies the planning model's complexity: from a planner's perspective, the modification of any link orientations does not impact on other relative joint angles, and therefore rotation actions can be sequenced \textit{in any order} the planner deems fit. 

Angles are specified using \textit{constants}, and are ordered using the predicate \texttt{OrientationOrd} as described by \eqref{eq:orientationord}. 
The difference between constant values is the \textit{granularity} of the resolution associated with modelled orientations.
For example, if \texttt{30} and \texttt{45} are used as constants representing, respectively, a $30$ and a $45$ $deg$ angle, then a predicate \texttt{(OrientationOrd 30 45)} is used to encode the fact that \texttt{30} precedes \texttt{45} in the orientation granularity, and corresponds to the description in \eqref{eq:orientationord_example}.    

Independently of what part of the articulated object is rotated, the domain model includes two actions, namely \texttt{RotateClockwise} (Figure \ref{fig:rotate_clockwise_basic}) and \texttt{RotateAntiClockwise}. 
Intuitively, the former can be used to increase the orientation of a given link of a certain granularity step (e.g., from $30$ to $45$ $deg$), whereas the latter is used to decrease the orientation, by operating on two connected links.
In the definition of \texttt{RotateClockwise}, \texttt{?l1} and \texttt{?l2} represent any two links $l_j$ and $l_{j+1}$, \texttt{?j1} is the joint $j_{j+1}$ connecting them, whereas \texttt{?o1} and \texttt{?o2} are the current and the obtained link orientations, respectively.
If \texttt{?j1} connected two \textit{different} links \texttt{?l1} and \texttt{?l2}, the angle \texttt{?o1} of \texttt{?l1} associated with \texttt{?j1} would be increased of a certain step (depending on the next orientation value) therefore leading to \texttt{?o2}.
A similar description can be provided for \texttt{RotateAntiClockwise}.

A problem is defined by specifying the initial and final states.
The former includes the \textit{topology} of the articulated object in terms of \texttt{Connected} predicates, and its initial configuration using \texttt{HasOrientation} predicates; the latter describes its goal configuration using relevant \texttt{HasOrientation} predicates.

\begin{figure}[t!]
\begin{mdframed}
\footnotesize
\texttt{
\begin{tabbing}
(:\=ac\=ti\=on\=\:\:\:Ro\=tateClockwise\\
\> :parameters (?l1 ?l2 - Link\\
\>\> ?j1 - Joint ?o1 ?o2 - Orientation)\\
\> :precondition (and \\
\>\> (Connected ?j1 ?l1)\\
\>\> (Connected ?j1 ?l2)\\
\>\> (not (= ?l1 ?l2))\\
\>\> (HasOrientation ?o1 ?j1)\\
\>\> (OrientationOrd ?o1 ?o2))\\
\> :effect \\
\>\> (and \\
\>\>\> (not (HasOrientation ?o1 ?j1)) \\
\>\>\> (OrientationOrd ?o2 ?j1)\\
\>\>\> (forall (?j2 - Joint ?o3 ?o4 - Orientation)\\
\>\>\>\> (when (and \\
\>\>\>\>\> (Affected ?j2 ?l1 ?j1)\\
\>\>\>\>\> (not (= ?j2 ?j1))\\
\>\>\>\>\> (HasOrientation ?o3 ?j2)\\
\>\>\>\>\> (OrientationOrd ?o3 ?o4))\\
\>\>\>\> (and\\
\>\>\>\>\>(not (HasOrientation ?o3 ?j2))\\
\>\>\>\>\> (HasOrientation ?o4 ?j2)))\\
\>\>)\\
)
\end{tabbing}
}
\end{mdframed}
\caption{The \textit{conditional} version of \texttt{RotateClockwise} in PDDL.}
\label{fig:rotate_clockwise_conditional}
\end{figure}
\textit{Absolute formulation}.
The absolute formulation differs from the relative one in that link orientations are expressed with respect to a unique, typically robot-centred, reference frame. 
Therefore, the set of link orientations is assumed to be directly observable by the robot perception system.
However, if a rotation action modifies a given link orientation $\theta^a_j$, all orientations of links in $up(l_j)$ or $down(l_j)$ must be consistently updated as well, i.e., it is necessary to propagate such change upstream or downstream.
As a consequence, such a representation increases the complexity of the planning task but it is more robust to errors: in fact, perceiving independent link orientations induces an \textit{upper bound} on the error associated with their inner angle.
The \texttt{Connected}, \texttt{HasOrientation} and \texttt{OrientationOrd} predicates are the same as in the relative formulation, subject to the different semantics associated with link orientations.
Also in the absolute formulation two actions are used, namely \texttt{RotateClockwise} (Figure \ref{fig:rotate_clockwise_conditional}) and \texttt{RotateAntiClockwise}.
However, with respect to the relative formulation, the effects of the action differ.
In particular, the model assumes that we can represent which joints are affected when a link is rotated around one of the corresponding joints.
This is done using the \texttt{Affected} predicate, i.e., a ternary predicate \texttt{(Affected ?j2 ?l1 ?j1)}, where \texttt{?l1} is the rotated link, \texttt{?j1} is the joint around which \texttt{?l1} rotates, and \texttt{?j2} is a joint affected by this rotation.
Therefore, if \texttt{?j2} were affected, the angle of the corresponding link would be modified as well in the conditional statement and, as such, it would affect other joints via the corresponding links. For each couple \texttt{?l1}, \texttt{?j1}, the list of joints affected by the corresponding movement should be provided under the form of multiple \texttt{Affected} predicates.
With reference to the action described in Figure \ref{fig:rotate_clockwise_conditional}, as in the previous case, the joint \texttt{?j1}, located between \texttt{?l1} and \texttt{?l2}, is increased by a quantity defined by a specific granularity, according to the \texttt{OrientationOrd} predicate. 
If rotating \texttt{?l2} around \texttt{?j1} affects \texttt{?j2}, the latter is updated, as well as all other joints upstream or downstream. This is encoded by the \texttt{forall} part of the PDDL encoding. Following the semantics of the language, the \texttt{forall} statement requires the planning engine to update the state of all joints \texttt{?j2} that are affected by the performed action -- checked conditions are specified via the \texttt{when} statement. The \texttt{HasOrientation} predicate of identified affected joints is then updated accordingly.
A similar definition for $\texttt{RotateAntiClockwise}$ can be easily given.

In terms of problem definition, beside \texttt{Connected} and \texttt{HasOrientation} predicates, it is necessary to include the list of appropriately defined \texttt{Affected} predicates.

It is noteworthy that the two action definitions, namely \texttt{RotateClockwise} and \texttt{RotateAntiClockwise}, are functionally equivalent.
Furthermore, any problem we target here could be solved -- in principle -- with just one action, as long as discretized angles were ring-connected.
We decided to introduce two different actions for two reasons:
on the one hand, it is rare that joints can rotate freely for $360$ $deg$ or more;
on the other hand, this model leads to shorter plans (on average) in terms of number of actions and cleaner, more natural executions, at the expense of a slightly longer planning time. 

\section{Experimental Validation and Discussion}
\label{sec:experimental_validation}

\subsection{System Implementation}
\label{sec:implementation}

\textsc{planHRC} has been implemented using both off-the-shelf components and \textit{ad hoc} solutions. 
All experiments have been carried out using a dual-arm Baxter manipulator.
The \textit{Perception} and \textit{Scene Analysis} modules are custom nodes developed using the Robot Operating System (ROS) framework.
They employ the Alvar tracker library to read QR codes\footnote{Webpage: \url{http://wiki.ros.org/ar\_track\_alvar}}.
Images are collected using the standard RGB camera of a Kinect device, which is mounted on the Baxter's \textit{head} and points downward to capture what happens on a table in front of the robot.  
\textit{Ontology} and \textit{Planning} have been implemented on top of ROSPlan \cite{Cashmoreetal2015}, which has been extended using the ARMOR framework for ontology management.
Two planners have been used, namely Probe \cite{LipovetzkyGeffner2011} and Madagascar \cite{Rintanen2014}.
The two planners have been selected on the basis of their performance in the agile track of the 2014 International Planning Competition, as well as following a computational assessment of their performance with respect to other planners with similar features\footnote{The interested reader can found relevant information in \cite{Capitanelli2017}.}.
The \textit{Execution} module and the various activated behaviours have been implemented using MoveIt!.

On-line, the architecture runs on a {8$\times$} Intel Core i7-4790 CPU $3.60$ GHz processors workstation, with 8 GB of RAM, running a Linux Ubuntu 14.04.5 LTS operating system.
Off-line performance tests about the planning process have been carried out on a workstation equipped with 2.5 GHz Intel Core 2 Quad processor, $4$ GB of RAM, running a Linux $2.6.32$ kernel operating system.

Problem formulations, as well as all generated instances, including domain, problems and plans, are freely available \footnote{Webpage: \url{https://github.com/EmaroLab/paco\_synthetic\_test}}.

\subsection{Planning Performance}
\label{sec:planning_perf}

Tests with synthetic problem instances have been performed to stress the two planning formulations.
For the tests, we varied the number of links $|L|$ from $4$ to $20$ and the number of allowed orientations $|O|$ a link can take from $4$ (i.e., with a resolution of $90$ $deg$) to $12$ (i.e., with a resolution of $30$ $deg$). 
As outlined above, such a resolution has a different meaning depending on whether we employ the absolute or relative formulations.

\begin{figure}[t!]
\centering
\includegraphics[width=100mm]{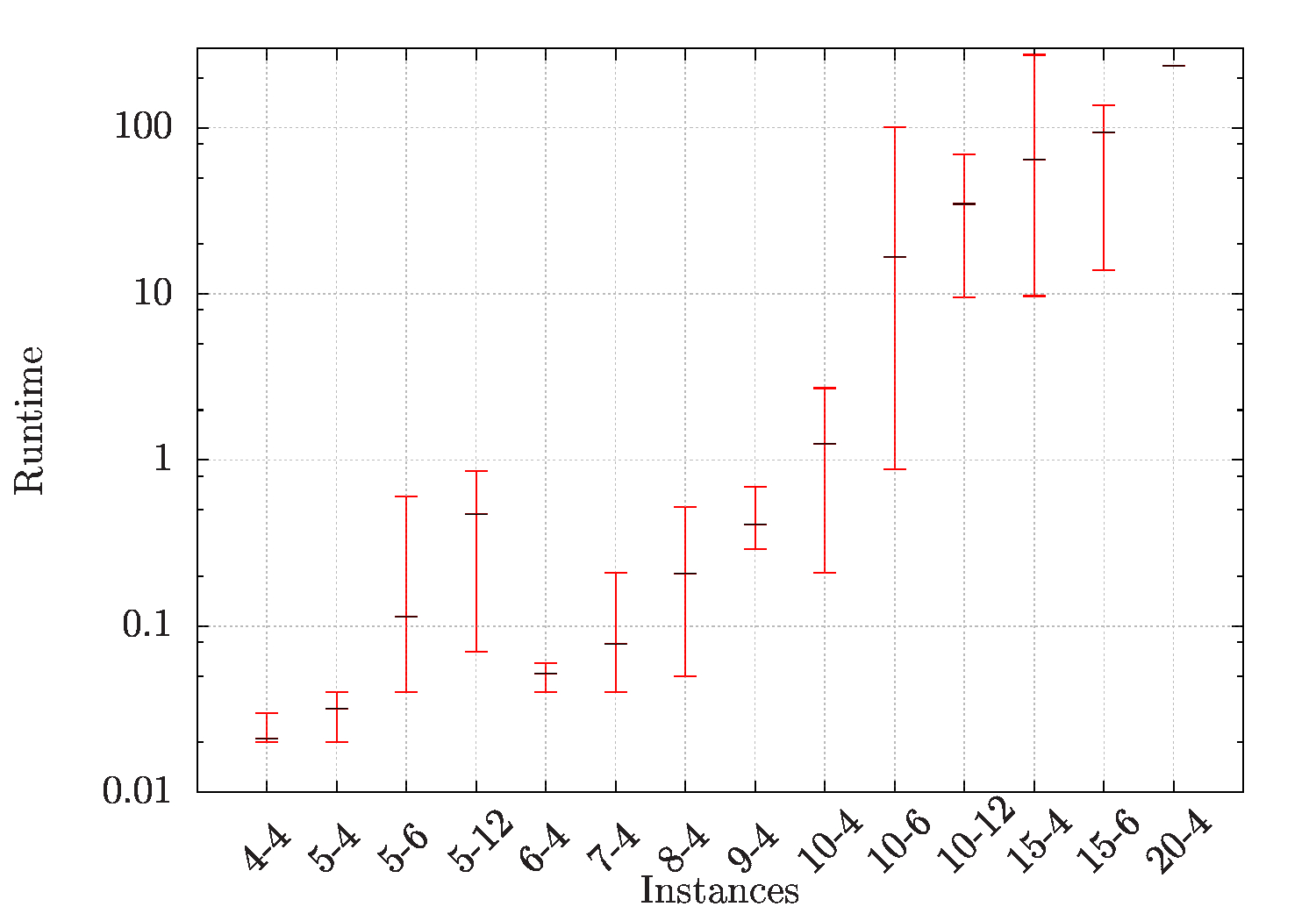}
\caption{Means and variances of solution times for different problem instances using the absolute formulation and Probe: on the x-axis, the first value indicates the number of links, the second the number of allowed orientations. Runtime is reported in $seconds$.}
\label{fig:absolute_probe}
\end{figure}
\begin{figure}[t!]
\centering
\includegraphics[width=100mm]{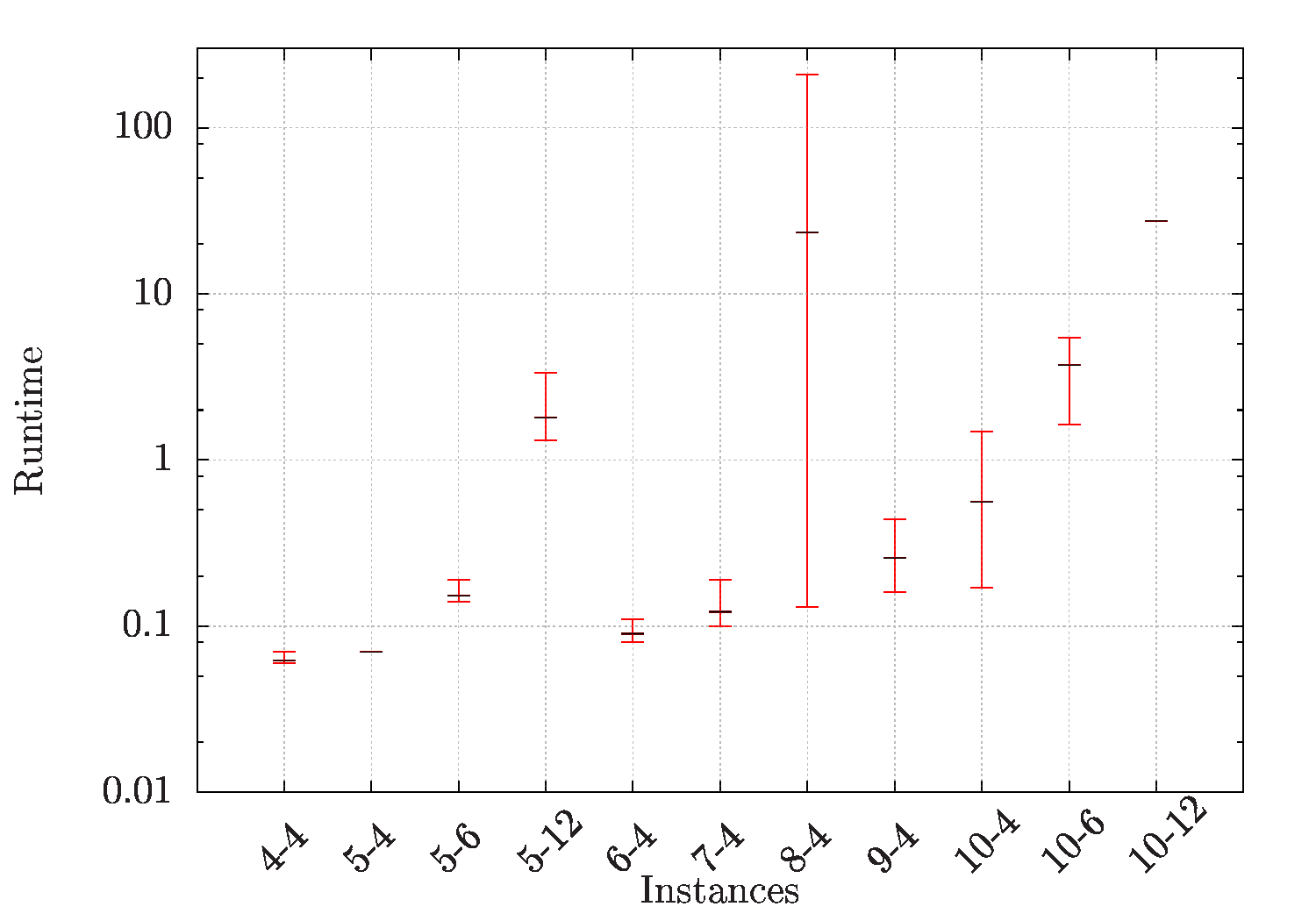}
\caption{Means and variances of solution times for different problem instances using the absolute formulation and Madagascar: on the x-axis, the first value indicates the number of links, the second the number of allowed orientations. Runtime is reported in $seconds$.}
\label{fig:absolute_mp}
\end{figure}
\begin{figure}[t!]
\centering
\includegraphics[width=100mm]{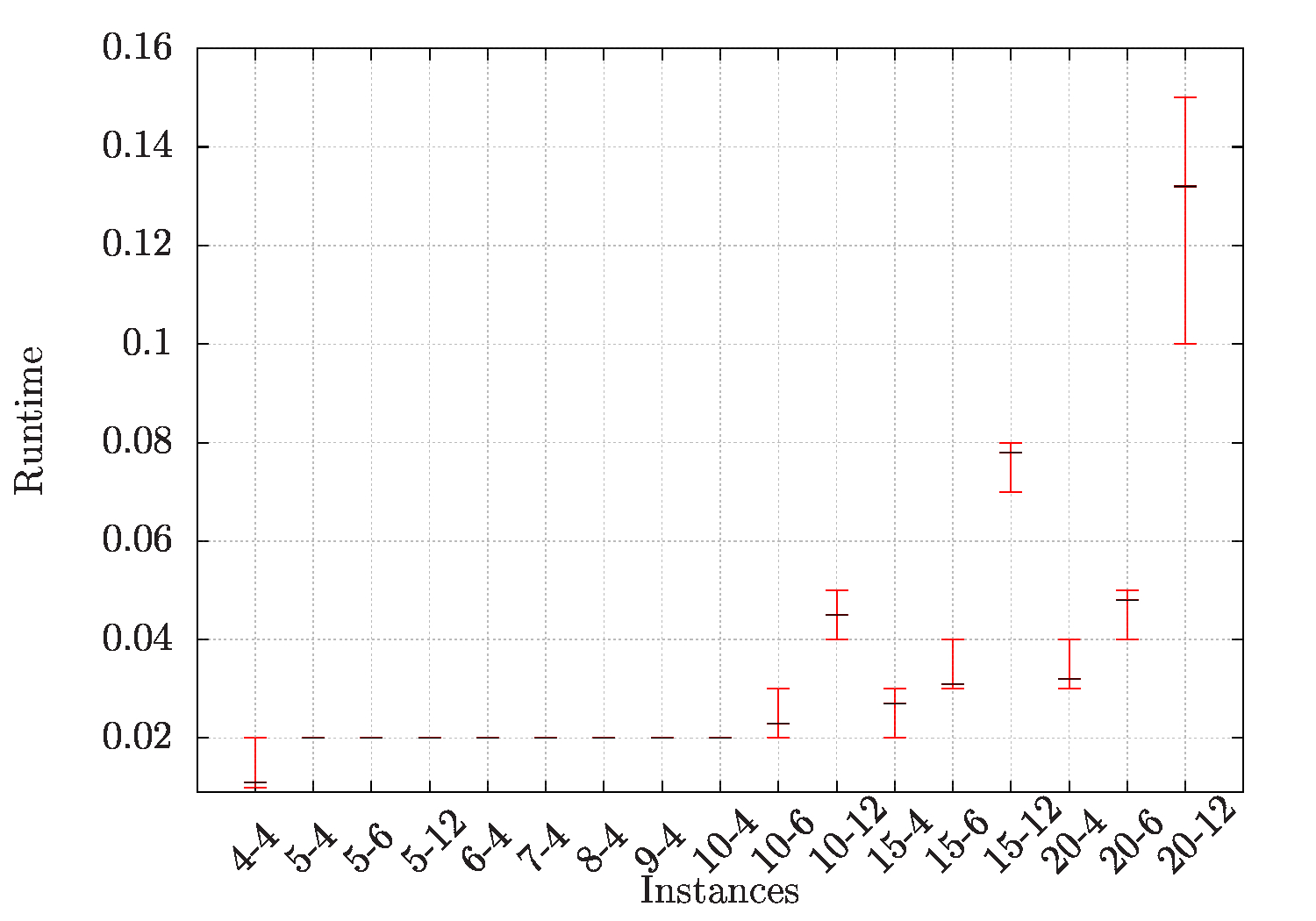}
\caption{Means and variances of solution times for different problem instances using the relative formulation and Probe: on the x-axis, the first value indicates the number of links, the second the number of allowed orientations. Runtime is reported in $seconds$.}
\label{fig:relative_probe}
\end{figure}
\begin{figure}[t!]
\centering
\includegraphics[width=100mm]{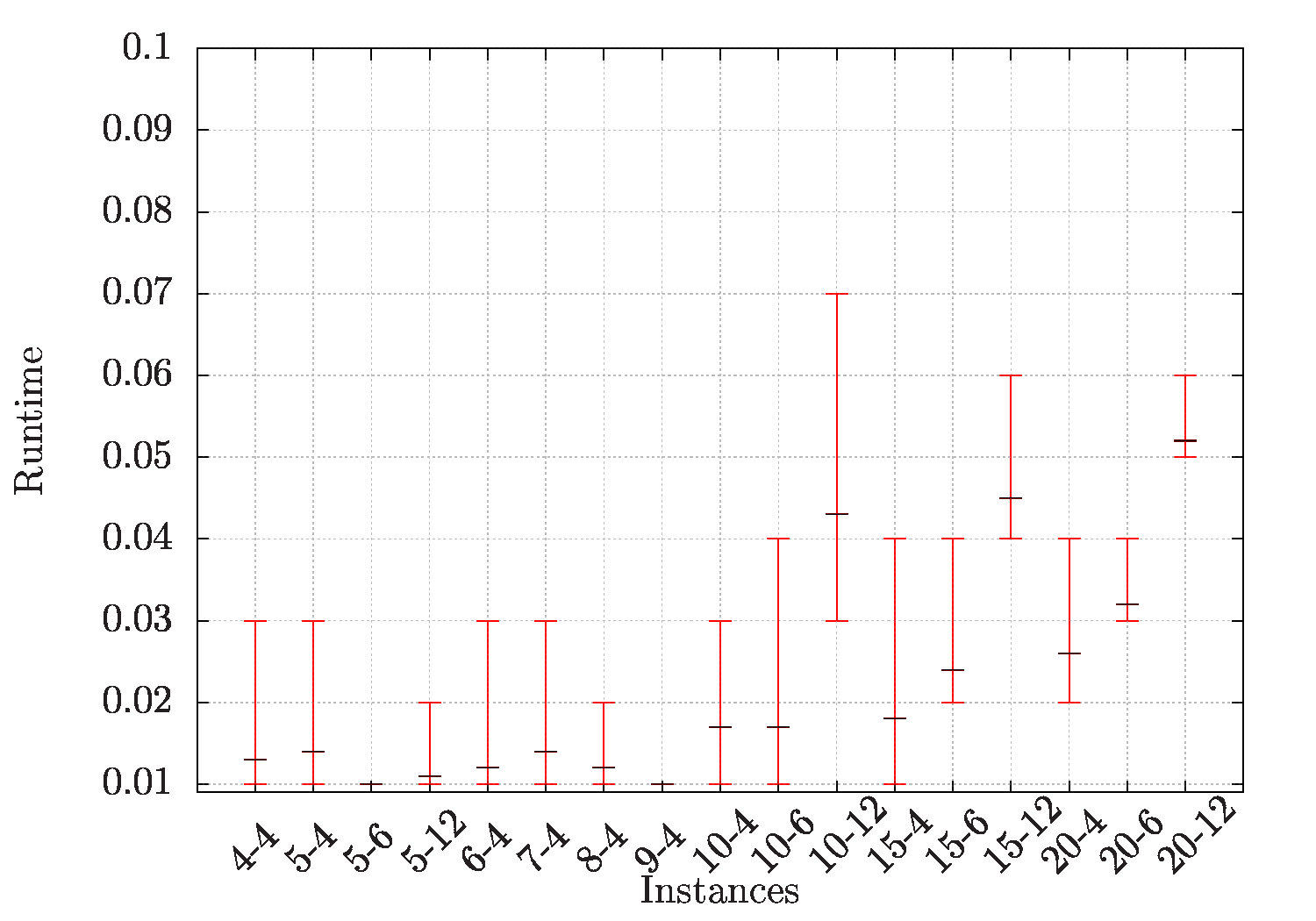}
\caption{Means and variances of solution times for different problem instances using the relative formulation and Madagascar: on the x-axis, the first value indicates the number of links, the second the number of allowed orientations. Runtime is reported in $seconds$.}
\label{fig:relative_mp}
\end{figure}
Figures \ref{fig:absolute_probe} to \ref{fig:relative_mp} show box plots representing means and variances, in $seconds$, for different problem instances, for all the combinations of formulation and planner.
Problem instances are labelled as ${x-y}$, where $x \leq |L|$ defines the number of links and $y \leq |O|$ specifies the orientation resolution. 
For each instance, planners have been executed $10$ times to take into account the randomness associated with the employed heuristics.
A $300$ $sec$ upper bound to the solution time has been set.
If a planner is unable to find a solution before such time limit is reached, it is stopped.
Figures only contain data related to problems solved within the time limit.

As it can be seen in Figure \ref{fig:absolute_probe}, when we use the absolute formulation and Probe, $73.5\%$ of the instances are solved, i.e., $125$ out of $170$.  
It is possible to observe that problem instances with up to $x \leq 10$ and $y \leq 4$ are solved in roughly less than $1$ $sec$, with a relatively small variance. 
When the number of links increase, planning time significantly increases as well, and thus the variance.
In the same situation, as depicted in Figure \ref{fig:absolute_mp}, Madagascar shows a more unpredictable behaviour: for small problem instances, it can quickly find a solution, and with a small temporal variance; however, the employed heuristics may cause large variances in specific cases, e.g., the instance labelled ${8-4}$. 
It is worthy to note that larger instances are rarely solved and, in general, the number of solved instances is lower when compared to Probe, i.e., only $53.5\%$ ($91$ out of $170$).
As it will be also showed in the next Section, these results seem to confirm hypothesis $H_1$, i.e., the more intuitive absolute formulation leads to more complex reasoning processes.
This is due to the fact that planners need to propagate the effects of each action to upstream or downstream links, which can be done only by employing a complex formulation involving conditional effects. 

If we consider the relative formulation, approach, then both Probe (Figure \ref{fig:relative_probe}) and Madagascar (Figure \ref{fig:relative_mp}) are very efficient, with Madagascar outperforming Probe to a small extent.
Both planners are capable of solving all the instances ($170$ out of $170$) in less that $0.2$ $sec$, and exhibit a very good scalability, as well as a very limited variance.
These results support hypothesis $H_3$, i.e., the reduced planning effort is reflected by the simpler formulation.

\begin{figure}[t!]
\centering
\includegraphics[width=100mm]{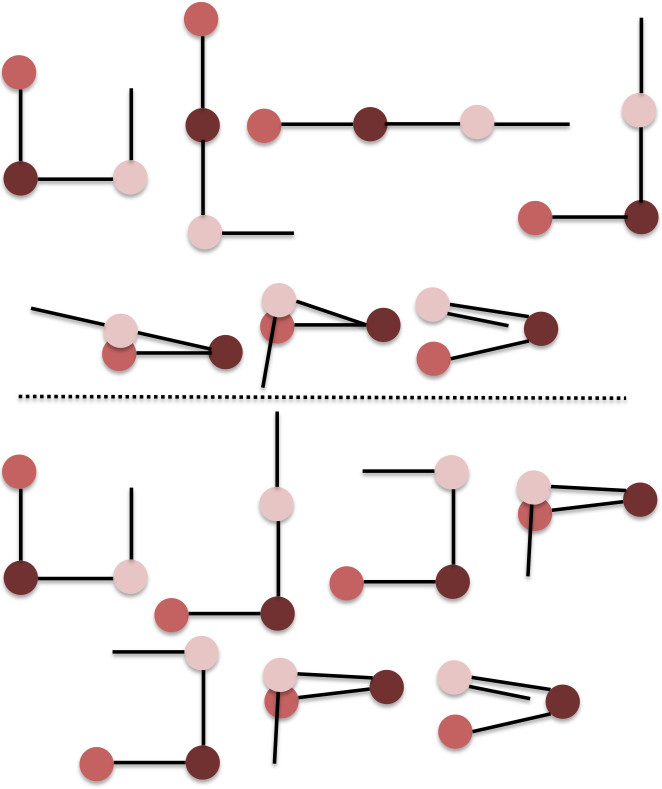}
\caption{A sequence of configurations for a ${3-3}$ problem, using the absolute formulation with Probe (first two rows) and Madagascar (second two rows).}
\label{fig:example_seq_absolute}
\end{figure}
\begin{figure}[t!]
\centering
\includegraphics[width=100mm]{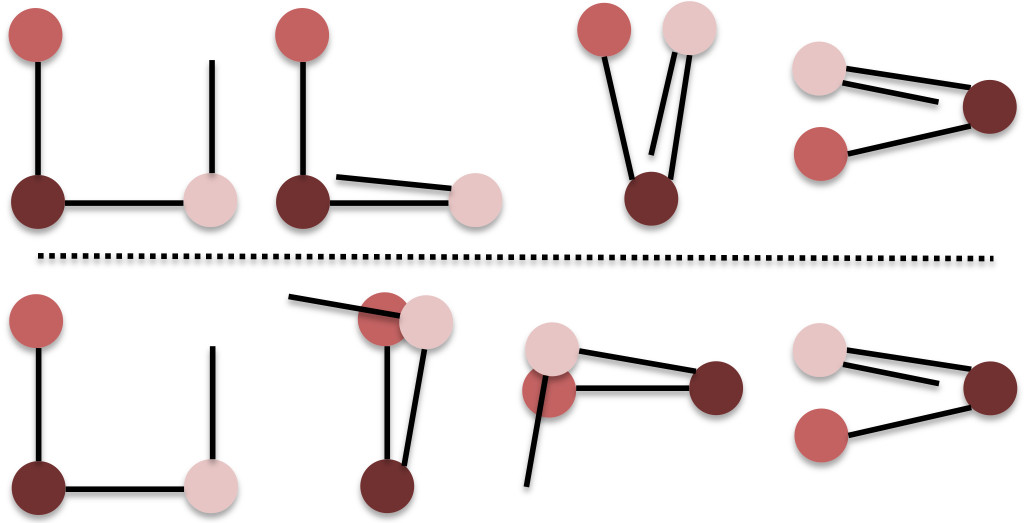}
\caption{A sequence of configurations for a ${3-3}$ problem, using the relative formulation with Probe (first row) and Madagascar (second row).}
\label{fig:example_seq_relative}
\end{figure}
\subsection{Examples}
\label{sec:examples}

In this Section, we provide examples of plans generated by Probe and Madagascar using the two formulations introduced above.
Furthermore, we show and discuss what happens in a number of human-robot cooperation use cases.

In order to discuss how the different planners deal with the absolute and the relative formulation, we focus the discussion on a specific instance with $3$ links and $3$ joints.
Figure \ref{fig:example_seq_absolute} shows two possible solutions, obtained respectively using Probe and Madagascar, when the absolute formulation is adopted.
The figure shows sequences of configurations assumed by the object, from left to right and from top to bottom.
It can be observed that plans are characterized by a number of seemingly unnecessary actions, since the planners must continuously maintain the representation consistency.
The plan obtained using Madagascar (on the bottom) also loops over two configurations, which is probably due to the employed heuristics.
This example seems to confirm $H_2$, i.e., the absolute approach leads to suboptimal plans, or plans which may not easily understood by human co-workers.

Figure \ref{fig:example_seq_relative} shows how Probe and Madagascar solve the same problem when a relative formulation is adopted.
Both planners generate solutions that are shorter than those obtained using the the absolute formulation, and no seemingly unnecessary actions are planned. 
In the plan generated by Madagascar, it is possible to observe that actions involving the same link tend to be performed sequentially, i.e., $H_4$ seems to be verified.
This holds for other solutions as well.

\begin{figure}[t!]
\centering
\includegraphics[width=100mm]{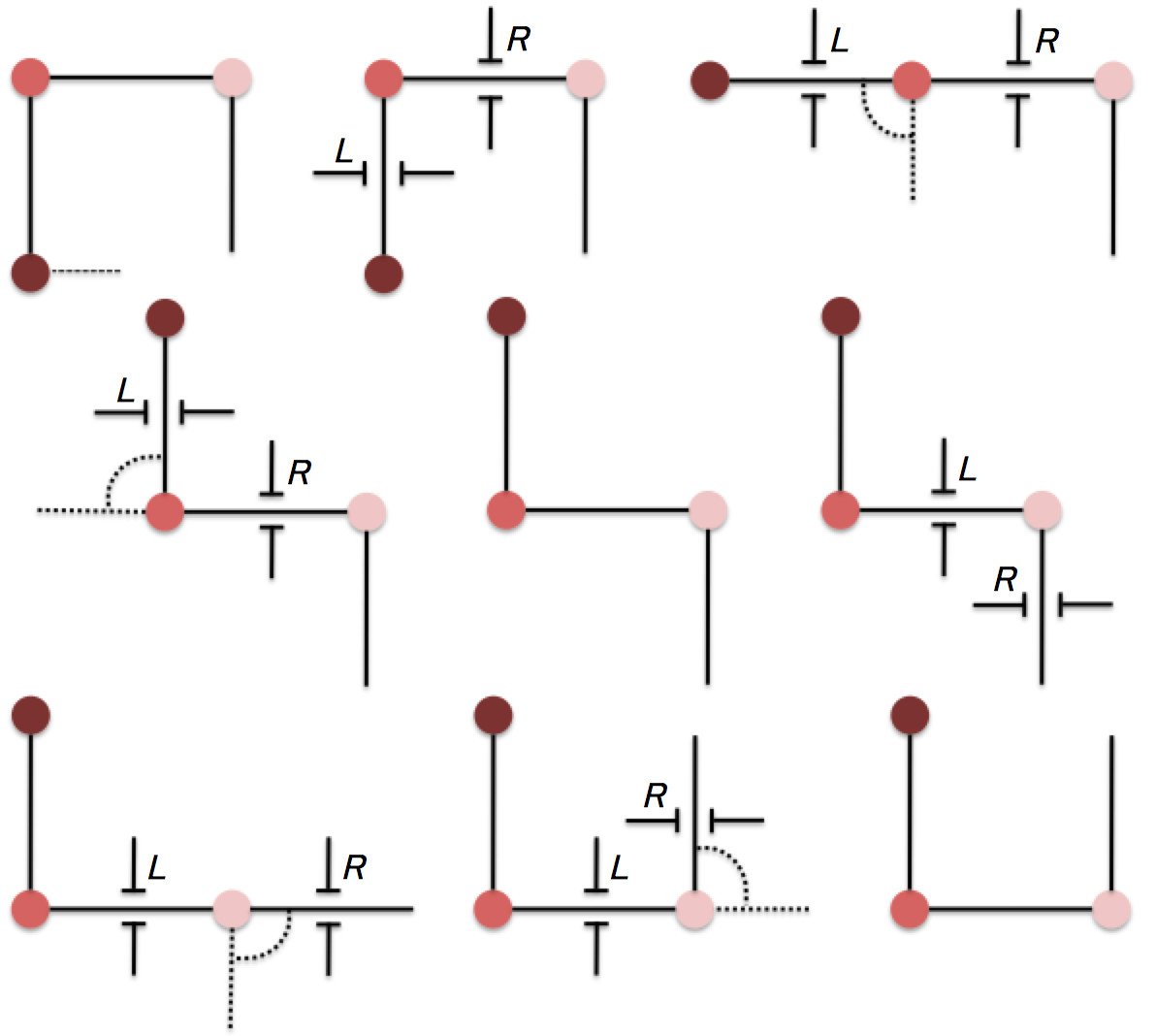}
\caption{A sequence of configurations for a ${3-4}$ problem, from top to bottom and left to right, as seen from the robot's perspective.}
\label{fig:sequence_conf}
\end{figure}
\begin{figure}[t!]
\centering
\includegraphics[width=120mm]{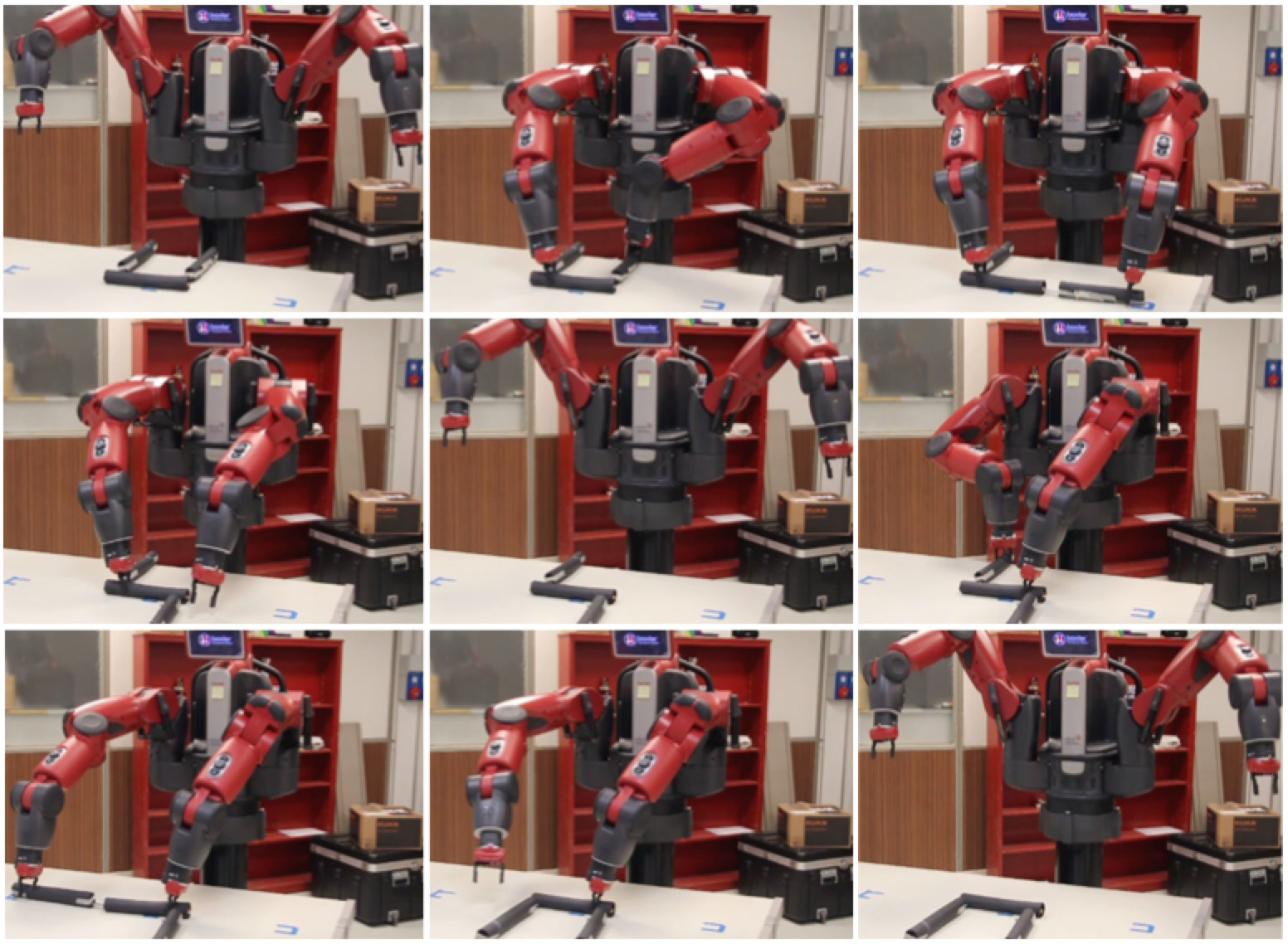}
\caption{The sequence of Figure \ref{fig:sequence_conf} as executed by Baxter without human intervention.}
\label{fig:sequence_baxter}
\end{figure}
As anticipated above, \textsc{planHRC} has been deployed on a dual-arm Baxter manipulator to enable the robot to autonomously manipulate articulated objects.
The Baxter operates on a $3$-link articulated object, assuming that the angle resolution is $90$ $deg$, i.e., a ${3-4}$ problem according to the definition introduced above.
Figure \ref{fig:sequence_conf} shows a sequence of configurations, including the initial and the goal ones, from top to bottom and left to right, whereas Figure \ref{fig:sequence_baxter} shows the corresponding relevant instants during the execution of the plan by the robot. 
It is worth noting that, each time a $\mathsf{RotateClockwise}$ or $\mathsf{RotateAntiClockwise}$ action is executed, the actual robot behaviour is made up of three steps:
the first is to firmly \textit{grasp} the link associated with the interested joint that must be kept still,
the second is to \textit{grasp} the link that must be rotated, the third is the actual \textit{rotation} of the proper amount.
In \textsc{planHRC}, this can be done indifferently by the left or right robot arms, according to a simple heuristics related to which arm is closer to the link to operate on.
Grasping actions in Figure \ref{fig:sequence_conf} are indicated with grasping signs close to the interested link, plus an $R$ sign to indicate that the action is performed with the right arm, or $L$ otherwise.
We decided not to model grasping actions at the planning level for two reasons:
on the one hand, they would have increased the burden of the planning process;
on the other hand, each rotation must be preceded by a grasping operation, and therefore this sequence can be easily serialized in the execution phase.

\begin{figure}[t!]
\centering
\includegraphics[width=120mm]{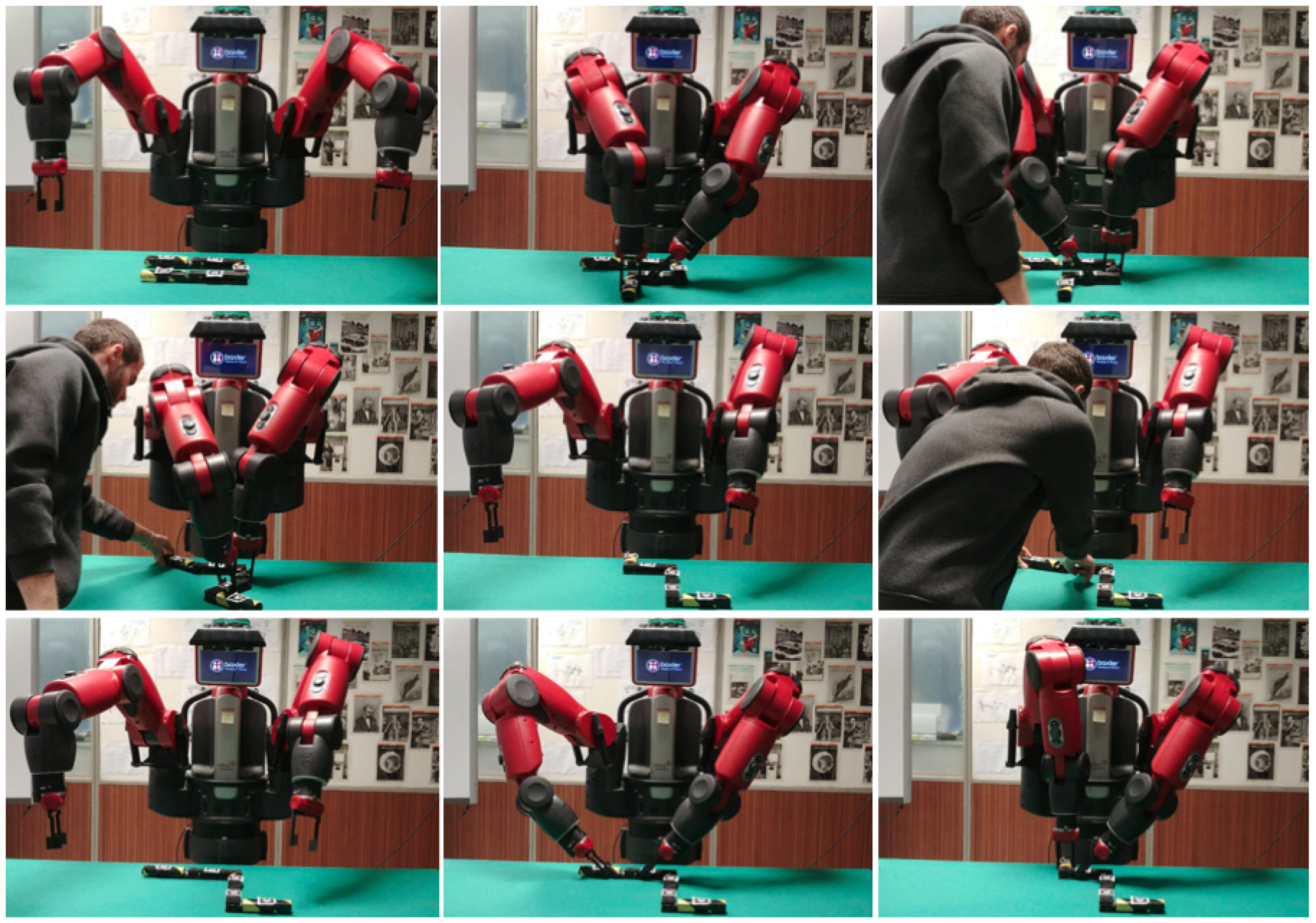}
\caption{A series of manipulation actions executed with the help of a human operator.}
\label{fig:seq_01}
\end{figure}
\begin{figure}[t!]
\centering
\includegraphics[width=120mm]{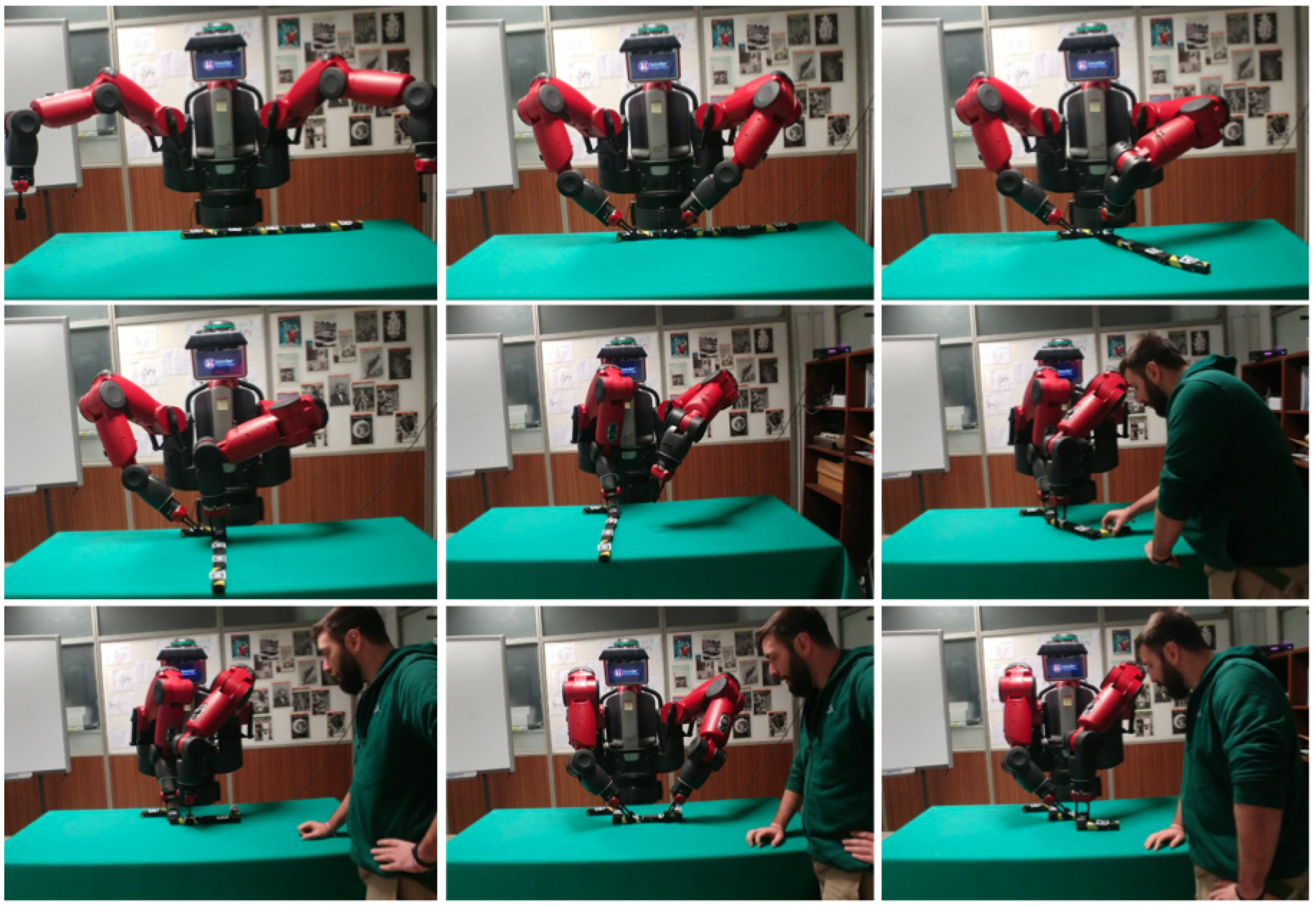}
\caption{Another series of manipulation actions executed with the help of a human operator.}
\label{fig:seq_02}
\end{figure}
Figure \ref{fig:seq_01} and Figure \ref{fig:seq_02} show two examples of plans where human intervention occurs to successfully accomplish the whole cooperation process.
In the figures, the two sequences must be analysed from top to bottom and left to right.

In Figure \ref{fig:seq_01}, it is possible to see that the human operator performs an action \textit{while} the robot is executing a rotation action on other links (snapshots $3$ and $4$).
The action performed by the human operator leads to a situation compatible with the object's target configuration.
As a consequence, the final configuration is reached in snapshot $5$. 
Afterwards, the operator modifies again the status of the first link (snapshot $6$), thereby leading to a configuration not compatible with the goal one. 
As a consequence, the robot intervenes to restore it (snapshots $8$ and $9$).
This sequence demonstrates two important features of \textsc{planHRC}:
first, the freedom human operators have in performing actions asynchronously with robot actions;
second, the robot capabilities in keeping the cooperation \textit{on track} coping with possible human mistakes. 

Figure \ref{fig:seq_02} shows an example where a human operator helps the robot complete an action, which was not performed in its entirety.
The robot starts executing a plan (snapshots $1$ to $5$).
However, a rotation action is not completed, leading the object's configuration to a state not compatible with the expected one (snapshot $6$).
Then, the human operator intervenes with an action aimed at completing the intended rotation and, at the same time, performing an additional rotation on the last link in the chain (snapshot $6$).
From that moment on, the robot autonomously completes the plan. 
This sequence shows how a plan can be successfully recovered by human intervention, and the fact that the robot can seamlessly continue plan execution.

\subsection{Discussion}
\label{sec:discussion}

On the basis of the requirements outlined above and the experimental analysis carried out to evaluate the whole \textsc{planHRC} architecture, it is possible to discuss a few interesting arguments, draw some conclusions, and indicate promising research directions.
In particular, the discussion that follows is focused on three aspects, namely planning speed, the generation of \textit{natural} sequences of manipulation actions and the resulting cooperation process when interacting with the robot.

\textit{Planning speed}.
The absolute and the relative formulations are characterized by different performance results. 

When using the absolute formulation, both Probe and Madagascar are capable of solving problem instances with a limited number of links and orientations in less than $1$ $sec$, which is a reasonable upper bound for the reasoning time of a collaborative robot interacting with a human co-worker, with Probe outperforming Madagascar on bigger problem instances.  
With around $10$ links, the time required to obtain a plan (if it exists) exponentially increases with the number of possible orientations, with solution times up to an average of $100$ $sec$ and beyond.
When using Probe, solution times for the same problem instance have a certain variance, which is almost uniform for different numbers of links and possible orientations. 
If Madagascar is used, such variance generally decreases, but sometimes it may become significantly large, as shown for example in the problem instance ${8-4}$. 
As far as human-robot cooperation processes are concerned, if an absolute formalization were used, then Probe would represent the best trade-off between complexity and solution times. 
In principle, Madagascar would be a better choice for problems with a reduced number of links and possible orientations, but the occasional presence of large variances in solution times would seriously jeopardize the human-robot cooperation process.  

The two planners behave differently when using a relative formulation.
Both Probe and Madagascar prove capable of solving large problem instances (i.e., with up to $20$ links and up to $12$ possible orientations) in less than $0.2$ $sec$. 
Solution times are exponential also in this case, but the very low time scale makes such trend relevant only to a limited extent.
Differently from the case with the absolute formulation, Probe behaves quite deterministically, and the same holds for Madagascar.
When dealing with human-robot cooperation, both planners are suitable to be used if a relative formulation is adopted, with a slight preference for Probe.

The relative formulation proves to be essential when the robot must deal with the directive $D_2$ discussed in the Introduction, and in particular to allow for a fast action re-planning when needed, as required by $R_3$. 

\textit{Natural action sequences}.
In general, the two formulations lead to qualitatively different plans, i.e., plans with different actions.

Independently of the employed planner, the absolute formulation originates plans, which are longer than those obtained using the relative formulation. 
In the first case, the solution may contain apparently unnecessary actions, as well as repeated sequences of actions.
This is due to the fact that when working on orientations of links located downstream in the chain, such orientations may be later modified as a side-effect when the algorithm operates on links upstream, therefore requiring reworking on downstream links.
Such plans are the result of certain planner heuristics.
However, they are often unnatural for humans to understand, which is of the utmost importance in human-robot cooperation processes.

Plans obtained starting from the relative formulation are shorter and -- in a generic sense -- more understandable by humans. 
Since the representation of orientations is relative for pairwise links, the planner does not need to modify orientations of downstream links multiple times, and solutions tend to include sequences of actions operating on the same link.
This makes plans easy to follow, irrespectively whether they are generated using Probe or Madagascar.

Thus, as far as naturalness is concerned, the relative formulation seems to be preferred over the absolute formulation. 
Shorter and easy-to-understand plans are supposed to strengthen a co-worker's ability to supervise robot actions in compliance with directive $D_2$ and to intervene when required, as prescribed by requirement $R_5$.
However, it is noteworthy that \textsc{planHRC} has not been tested with human subjects yet.
As a consequence, there are still to-be-validated hypotheses requiring us to conduct a specifically designed study, also related to the role of context-aware planning in human-robot cooperation \cite{Mastrogiovannietal2013}.

\textit{The cooperation process}.
On the one hand, in absence of errors related to action execution, once a plan is available \textsc{planHRC} should be able to carry it out in its entirety.
This is in agreement with directive $D_1$ discussed in the Introduction.
On the other hand, when either one action is not executed successfully or it has been carried out only partially, a human co-worker can intervene to obtain an object configuration that the robot can operate upon.
As a whole, these two facts support requirement $R_2$.

As described above, before any action is executed, the robot checks whether a number of expected normative predicates hold in the current planning state. 
Implicitly, this means that any error in action execution or human intervention is synchronously assessed before the next planned action can start.
Obviously enough, this represents a limiting factor for \textsc{planHRC}, and originates from the focus on planning \textit{states} rather than actions. 
A more flexible reactive system may make use of human actions to determine causes of faults on the fly, instead of being limited in assessing their outcomes at discrete intervals. 
However, it also enforces the fact that humans are in control at any time: the robot simply waits for human intervention to finish and then plans a course of action from that moment on.  

Also in this case, such an approach should be validated with human subjects. 
Current work is devoted to investigate these matters.  

\section{Conclusions}
\label{sec:conclusions}

The paper introduces a hybrid reactive/deliberative architecture for collaborative robots in industrial scenarios.

We focus on the collaborative manipulation of flexible or articulated objects.
We consider articulated objects as suitable models for flexible objects, and a number of challenges are considered:
(i) the representation of articulated objects using a standard OWL-DL formalism, and the definition of suitable planning models using PDDL; 
(ii) an implicit addressing of human operator's actions; and and
(iii) an assessment of how perception assumptions and the related representation schemes impact on planning and execution.

The developed architecture is evaluated on the basis of a number of functional and non functional requirements: the possibility for the system to implicitly recognize the effects of human actions, the robot's capabilities in adapting to those actions, a fast (re-)planning process when needed, just to name the most important ones. 

Current work is focused on two aspects:
on the one hand, a more detailed, computationally efficient, representation of articulated objects and the corresponding planning models; on the other hand, a systematic evaluation of the architecture with human volunteers.

\section*{References}
\bibliography{references}

\end{document}